\documentclass[10pt,a4paper,journal,cspaper,compsoc]{IEEEtran}
\ifCLASSOPTIONcompsoc
  \usepackage[nocompress]{cite}
\else
  \usepackage{cite}
\fi
\ifCLASSINFOpdf
  \usepackage[pdftex]{graphicx}
\else
\fi
\usepackage[cmex10]{amsmath}
\usepackage{algorithmic}
\usepackage{algorithm}

\usepackage{array}

\usepackage{mdwmath}
\usepackage{mdwtab}
\ifCLASSOPTIONcompsoc
  \usepackage[caption=false,font=normalsize,labelfont=sf,textfont=sf]{subfig}
\else
  \usepackage[caption=false,font=footnotesize]{subfig}
\fi
\usepackage{stfloats}

\ifCLASSOPTIONcaptionsoff
  \usepackage[nomarkers]{endfloat}
 \let\MYoriglatexcaption\caption
 \renewcommand{\caption}[2][\relax]{\MYoriglatexcaption[#2]{#2}}
\fi
\newtheorem{definition}{Definition}

\begin{document}
%
\title{Optimization in Differentiable Manifolds in Order to Determine the Method of Construction of Prehistoric Wall-Paintings}
%
%
%
%

\author{Dimitris~Arabadjis, Panayiotis Rousopoulos, Constantin Papaodysseus, Michalis Exarhos, Michalis Panagopoulos,~\IEEEmembership{Member,~IEEE,} and Lena Papazoglou-Manioudaki
\IEEEcompsocitemizethanks{\IEEEcompsocthanksitem D. Arabadjis, P.Rousopoulos, Constantin Papapodysseus and Michalis Exarhos are with the School of Electrical and Computer Engineering, National Technical University of Athens, Iroon Polytechniou 9, 15773, Athens, Greece.
E-mail: cpapaod@cs.ntua.gr
\IEEEcompsocthanksitem M. Panagopoulos is with Department of Audio \& Visual Arts, Ionian University, Corfu, Greece.
\IEEEcompsocthanksitem L. Papazoglou-Manioudaki is with National Archaeological Museum of Greece, Patision 44, Athens}
\thanks{\copyright ~ 2011 IEEE. Personal use of this material is permitted. Permission from IEEE must be obtained for all other uses, in any current or future media, including reprinting/republishing this material for advertising or promotional purposes, creating new collective works, for resale or redistribution to servers or lists, or reuse of any copyrighted component of this work in other works.}
}

\IEEEcompsoctitleabstractindextext{%
\begin{abstract}
In this paper a general methodology is introduced for the determination of potential prototype curves used for the drawing of prehistoric wall-paintings. The approach includes a) preprocessing of the wall-paintings contours to properly partition them, according to their curvature, b) choice of prototype curves families, c) analysis and optimization in 4-manifold for a first estimation of the form of these prototypes, d) clustering of the contour parts and the prototypes, to determine a minimal number of potential guides, e) further optimization in 4-manifold, applied to each cluster separately, in order to determine the exact functional form of the potential guides, together with the corresponding drawn contour parts. The introduced methodology simultaneously deals with two problems: a) the arbitrariness in data-points orientation and b) the determination of one proper form for a prototype curve that optimally fits the corresponding contour data. Arbitrariness in orientation has been dealt with a novel curvature based error, while the proper forms of curve prototypes have been exhaustively determined by embedding curvature deformations of the prototypes into 4-manifolds. Application of this methodology to celebrated wall-paintings excavated at Tyrins, Greece and the Greek island of Thera, manifests it is highly probable that these wall-paintings had been drawn by means of geometric guides that correspond to linear spirals and hyperbolae. These geometric forms fit the drawings' lines with an exceptionally low average error, less than 0.39mm. Hence, the approach suggests the existence of accurate realizations of complicated geometric entities, more than 1000 years before their axiomatic formulation in Classical Ages.
\end{abstract}

}

\maketitle

\IEEEdisplaynotcompsoctitleabstractindextext

%
\IEEEpeerreviewmaketitle

\section{Introduction}
\label{intro}
In this paper a methodology of general applicability is presented for answering the question if an artist used a number of archetypes to draw a painting or if he drew it by free hand. In fact, the contour line parts of the drawn objects that potentially correspond to archetypes are initially spotted. Subsequently, the exact form of these archetypes and their appearance throughout the painting is determined. The method has been applied to celebrated wall paintings, "Lady of Mycenae", of the 13th century B.C. excavated at Mycenae, Greece and "Naked Boys" drawn c. 1650 B.C. excavated at Akrotiri, Thera, Greece.

\subsection{Previous Works on Fitting Data to Curve Prototypes}
\label{sec:99}
In the bibliography there are various approaches to the problem of optimally fitting curves to 2D-data, either using the explicit or the implicit form of the model curves. Namely, in \cite{OrthDistFit}, the optimal fit of implicit plane curves to data points is treated via a square distance minimization procedure. This minimization is performed over translation, rotation and implicit function parameters, in order to deal with rigid body motion of the model curve and with its shape variance. Previously, in \cite{LSodist}, authors had employed orthogonal distances and the related tangential quantities between data and model curve points, in order to optimally fit conic sections to the given data points, via a non-linear regression algorithm. A lower bound for the least squares non-linear regression, between an implicit model curve and the data points of a contour, is given in \cite{LowBound}, thus offering a measure for the optimality of the orthogonal distance curve fitting methods. In \cite{PapFrag}, authors analytically determine the least-squares optimal translation and rotation of an explicit curve model, in order to fit a given set of data points. Then the primary parameters of the model curve are obtained via a 2D iterative region descending process. The alignment of an implicit curve model to given contour points is also dealt in \cite{AffFit}, in a manner that the fitting process is invariant under affine transformations. Namely, the authors use affine invariant Fourier descriptors for an area-parametrization of the model curve and, by matrix annihilation, they determine the harmonic implicit form for the model curve descriptive equation. In \cite{new0} authors exploit point-wise distribution of shape templates in order to define shape descriptors and develop the corresponding distance measure. This distance
measure offers exact point-wise correspondence between two different shapes, allowing for optimal matching transformation of one shape to the other. In \cite{new3} authors deal with the problem of fitting explicitly described B-Spline model curves to data-points lying on a 2D-surface. The B-Spline is restricted so as to lie on the same surface with the data-points by treating the surface as a differentiable 2-manifold. Then, the fitting procedure is performed so as the on manifold geodesic distance between data and model curve points is minimized. Analogous technical elements for modifying planar objects so as to lie on a manifold are developed in \cite{new1}, where the authors re-evaluate the "Beltrami short time kernel" in the case that the image to be filtered lies on a manifold. In \cite{Taubin2} authors deal with the problem of fitting algebraic curves to data points, via their implicit representation, by constraining their functional form so as to have closed zero level sets, thus making the curve determination more robust. The implicit functional form of the model curves (algebraic) is adopted in \cite{3LImp} to define itself a matching error via the polynomial function level sets. In order to deal with arbitrary orientation, authors adopted invariant computation techniques of polynomial coefficients. Finally, in \cite{Taubin1} the representation of an object as a unification of sub-objects, that can be analytically described by implicit curve or surface forms, is used, in order to segment a complex visual form into "meaningful" sub-shapes.

\subsection{The Present Problem and the Proposed Approach}
In the present work, the demand was the determination of an unambiguous relative placement between a prototype implicit curve and the contour data, in the sense that attribution of various contours data to a prototype curve is orientation invariant. Namely, given an implicit form of a prototype curve, the integral of the least distances between a curve part and each contour data points is computed immediately, by means of the flat curvature of the implicit functional form of the prototype curve, with no need of intermediate optimal placement of the data points along the prototype curve points. Additionally, for the present application, one must spot a unique prototype curve that gave rise to a considerable number of different contour realizations, where each such realization corresponds to a different part of the prototype curve. So, the determination of the model curve that optimally fits a set of partial realizations of its planar graph is a problem of increased degrees of freedom, compared with the widely treated problems of curve fitting. Namely, the problem of optimally fitting the model curve to the data runs over the space of the free parameters of the curve model and, simultaneously, over all possible line segments of it and all possible data points subsets that could have been generated by the same prototype. There are various approaches that deal with the determination of the parameters that optimally fit a model curve to a given set of points in a statistically efficient manner. In \cite{Statist}, there is a compilation of the existing implicit curve fitting techniques, together with statistical analysis of their efficiency. Concerning the works referred and outlined in Sect. \ref{sec:99}, they are mainly split into 2 categories: a)those who treat the problem of fitting curve models to data points and b) those who treat the problem of shape matching. Among these publication there are approaches that treat the problem of curve fitting (e.g.\cite{OrthDistFit}, \cite{AffFit}, \cite{PapFrag}, \cite{3LImp}), or shape matching (e.g. \cite{new0}) with an orientation invariant approach, but none of these techniques dealt with the problem of simultaneous fitting the same implicit curve model to different sets of data points. Moreover, in most curve-fitting approaches, determination of the optimal values for the free parameters  of the model curve is based on iterative error minimization algorithms like Gauss-Newton, Gradient Descent, etc. that do not necessarily converge to a global error minimum. Concerning shape matching methods, although they offer unsupervised shape similarity measures, minimization of these measures deforms the model shape without constraining any of its geometrical characteristics.
In order to deal with the demands of unambiguously optimal fit between the determined model curve and all contour data attributed to it, we had to develop a unified optimization method of the least possible dimensionality that exhaustively acts on the set of all possible prototypes of the same parametrized curves family. The fitting process introduced here treats each class of prototype curves as a 4-manifold and performs a curvature driven exhaustive error minimization between the class of prototype curves and the available data points. By applying this methodology two times, first with respect to the drawn contours and next with respect to the prototypes (Sects. \ref{sec:5_1}, \ref{sec:7} respectively), we determine a unique prototype curve that optimally fits all data points attributed to it.

\section{Fundamental Notions and Definitions Concerning the Method of Wall-Paintings Drawing}
\label{sec:1}

\textsl{A first crucial hypothesis for the method the artist(s) might employed for drawing the wall-paintings c. 3500 years ago}\\
"The fresco technique asked for highly fast and precise execution of the drawing process. At the same time, the stability of the contour line of the figures depicted on the painting had always been a primary goal of the artists throughout human history. Thus, a plausible assumption about the method of drawing of these prehistoric wall paintings is that the artists employed guides or instruments to support the drawing object." The quality of drawing of the paintings considered in this paper, suggests an additional content to this assumption: "the artists took special care and paid attention to ensure continuity of the drawn contour line and, wherever possible, of its tangent, at the points where a change of guide occurred."

\begin{definition}[The object part]
If the aforementioned assumption is correct, then there will be subsets of the contour line of the various figures appearing in the wall-painting, which they had been drawn by a continuous stroke of the paintbrush. These subsets are called object parts. 
\end{definition}
\begin{definition}[The object]
One can define the object to be a subset of the wall painting border, which represents a thematic unit, is smooth and its beginning and end points are discontinuities of the border line or of its tangent. Evidently an object is a concatenation of object parts, since it has been drawn by a usually small number of continuous strokes of the paintbrush.
\end{definition}

\subsection{A rigorous determination of the object part}
\label{sec:2}
Consider ideally that, in the wall painting, all objects' contour lines are continuous, described by the piecewise twice differentiable mono-parametric vector equation $\vec r(t)=(x(t) , y(t))$.
Then, an object part is spotted by determining a contiguous subset of the object curve whose beginning and end points are one of the following: 1)the beginning or the end of the object, 2)a point where the curvature is discontinuous. Analogous pictorial feature definitions can be found in \cite{new2}, where authors use curvature extremes and sign changes in order to define strokes, corners, endpoints and junctions.
	
	The notion of an object part is crucial, since if the artist(s) indeed used stencils, then, each time he placed the stencil on the wall and drew a line, we consider he created an object part.

	If $L_O$ is the length of an arbitrary object then we select an appropriate small percentage of $L_O$, say $L_S$. Subsequently, in order to spot points where there is change of stencil, we proceed as follows: Let an arbitrary point on object contour, say the i-th, where $L_S \leq i \leq L_O - L_S$, then we determine the 3rd degree polynomials $\vec r _i^3 \left(x_i^3(s) , y_i^3(s) \right)$ of object length $s \in [0,L_S]$ that best fit object points $\vec p (i-L_S) ... \vec p (i)$ and the polynomial of 5th degree $\vec r _i^5 \left(x_i^5(s) , y_i^5(s) \right)$ of object length $s \in [0,L_S]$ that best fits the points $\vec p (i) ... \vec p (i+L_S)$ and satisfies the tangent continuity on $\vec p (i)$, namely $\frac{d}{ds} \vec r _i^3(L_S) = \frac{d}{ds} \vec r _i^5(0) = \left(\dot x_i^c , \dot y_i^c \right)$.
	 
	In order to decide if the i-th pixel is an ending point of an object part we check for discontinuity of the object contour curvature on it. Hence, the two approximations of the contour curvature from the left $c_i^- (L_S)=\ddot{\vec r}_i^3(L_S)$ 
and from the right $c_i^+ (0) = \ddot{\vec r}_i^5(0)$ 
should manifest an abrupt difference. Equivalently, we demand $\left| c_i^-(L_S) - c_i^+(0) \right| \geq \delta \theta$, where $\delta \theta$ a very small angular threshold. If this demand is met then the i-th point $\vec p (i)$ is the end of the current potential object part. The beginning of the next potential object part is the first point after i-th, say the j-th, where the demand $\left| c_i^-(L_S) - c_i^+(0) \right| \leq \delta \theta$ is met.
	By application of this method to all available objects of the wall-painting figures, we generate an ensemble of potential object parts with contour point vectors $\vec r _p$, $p=1,...,M_{op}$. We once more stress that we consider that each such object part was drawn by one paintbrush stroke. We also assume that if we achieve in corresponding these object parts into a small number of prototypes, then we will support the hypothesis that these "one-stroke contour lines were made by means of a stencil or another equivalent instrument that guided the brush." If we further adopt the plausible assumption that each such stencil had a specific geometric form, namely in modern mathematics a specific functional type, then we may deduce that the derivatives of all orders of $\vec r (t)$ remain continuous along an object part. On the contrary, at the points of change of stencil, or of replacement of the stencil, it is highly probable that the curvature will essentially change, even in the case where the artist(s) achieve continuity of $\vec r (t)$ and $\dot {\vec r} (t) $ to ensure a smooth and good aesthetic result.

\section{A Brief Description of the Employed Methodology for Determining Possible Geometric Guides}
\label{sec:3_p}

\begin{table*}
\caption{Basic Symbols}
\label{tab:5}       
\begin{tabular}{ll}
\hline\noalign{\smallskip}
Symbol & Meaning  \\
\noalign{\smallskip}\hline\noalign{\smallskip}
$f_S(\chi,\alpha) = 0$ & implicit description of a model curve \\
$\chi$ & $(x,y)$ Cartesian coordinates \\ 
$\alpha$ & $(a,b)$ primary parameters of the model curve \\
$c_S(\chi,\alpha)$ & the flat curvature function of the $S$ model curve $c_S=\nabla_{\chi} \cdot \frac{\nabla_{\chi}f_S}{\left\| \nabla_{\chi}f_S \right\|}$\\
$\chi_S(\xi,\alpha)$ & the model curve, parametrized via curve length $\xi$, which is resulted by a fixed pair $\alpha$ of primary parameter values \\
$\alpha_S(\beta, \chi) $ & the primary parameters isocontour, parametrized via curve length $\beta$, as it is resulted by\\ & a fixed pair $\chi$ of Cartesian planar coordinates \\
$\delta_{\chi}$ & the curve length while moving on $xy$-plane in a direction normal to $\chi_S$ \\
$\delta_{\alpha}$ & the curve length while moving on $ab$-plane in a direction normal to $\alpha_S$\\
$c_{\chi}^{\alpha}$ & the internal product between the directions normal to $\alpha_S$ and $\chi_S$\\
$\left( T_{\chi} , T_{\alpha} \right)$ & $\left(\left\| \partial_{\chi}f_S \right\| , \left\| \partial_{\alpha}f_S \right\| \right)$\\
$\left(d \eta, d \gamma, dr_{\chi}^{\alpha}, d \gamma_{\xi}^{\beta} \right)$ & The local frame defined by the level-sets of $f_S$: $d\eta$ is the curve length differential along $\nabla f_S$ and\\ & the remaining curve length differentials evolve along the curves selected to define the tangent space of $f_S$:\\ & $d \gamma$ evolves along $(\chi_S,\alpha_S)$ direction, $d r_{\chi}^{\alpha}$ evolves in a direction normal to covariance of $\chi$ and $\alpha$ \\ & and $d \gamma_{\xi}^{\beta}$ normal to covariance of $\chi$ and $\beta$ \\
$(\hat n , \hat l)$ & the unit directional vectors normal and tangent to $\chi_S$ respectively\\
$(\vec n_S , \vec l_S)$ & the unit directional vector spaces normal and tangent to the level-sets of $f_S$ respectively\\
$K_S(\chi,\alpha)$ & the Euclidean norm of $\vec n_S$ evolution on the level-sets of $f_S$ (i.e. of the level-sets' curvature)\\
\noalign{\smallskip}\hline
\end{tabular}
\end{table*}

After application of the aforementioned methodology, we have determined an ensemble of object parts, i.e. an ensemble of parts of the painted figures' borders with small curvature fluctuations along each such part. Now, the question arises if there is a small number of prototype curves that optimally fit these object parts and if yes, how to determine the exact functional form of these prototypes. In order to deal with this problem we have applied a novel methodology outlined in the following steps. \\
\textsl{Step  1 - Choice of a reasonable criterion that quantifies how well an object part fits a prototype curve part} \\
Suppose the prototype curve $S$ in $\Re^2$ that is described by the implicit functional form $f_S(x,y)=0$ and let $\hat n$ be the unit vectors normal to it. Moreover, let an arbitrary point $M$ on an object part and its vectorial distance $\vec \delta_M$ from $S$. Suppose, for a moment, that the object part is a continuous curve. Then, evidently, a measure for the level of fitting the considered object part to $S$ is the integral of $\delta_M = \left\| \vec \delta_M \right\|$ along a proper part of $S$.\\
\textsl{Step  2 - Alternative version of the aforementioned criterion}\\
Because $\vec \delta_M$ is calculated along $\vec n_S$, by applying Stokes theorem on domain $\Omega$, bounded by the object part and the curve part of $S$, we obtain.
\begin{equation}
2 \int_S \vec \delta_M \cdot \hat n dl = \int_{\partial \Omega}\delta_M dl = \int_{\Omega}(\nabla \cdot \hat n) d \Omega
\end{equation}
But, actually, $\nabla \cdot \hat n = c_S$, where $c_S$ is the flat curvature of the implicit functional form $f_S$ of the curve model $S$. Consequently, minimization of the integral of $\delta_M$ on a proper part of $S$ can be obtained by minimization of the corresponding area integral of $c_S$. In Sect. \ref{sec:4} and in appendix A, we show that minimization of the integral of the flat curvature over $\Omega$ can be obtained by minimizing the integral of $c_S$ on the boundary of $\Omega$, $\partial \Omega$ \\
\textsl{Step  3 - Primary Parameters definition}\\
In general, the functional form of $f_S$ depends on a significant number of parameters. For example, conic sections implicit functional form depends on 5 parameters in general. In Sect. \ref{sec:5_0}, it is shown that, independently of the number of parameters and the exact functional form of $f_S$, two parameters are sufficient to describe all possible deformations of the flat curvature of $f_S$ on the $(x,y)$-plane. We will call these parameters as the "primary parameters" of $f_S$.\\
\textsl{Step 4 - Extending the fitting process in $\Re^4$}\\
Assume a given object part, OP and a parametrized class of prototypes $S$, implicitly described by the equation $f_S(x,y,a,b) = 0$. We look for the determination of the proper primary parameters values $(a,b)$ and the proper relative placement between OP and $S$, so as the integral of the point by point distances (see Step 1) between these curves are minimized. Hence, we extend the object part - prototype fitting analysis from $\Re^2$ to $\Re^4$, where two of the coordinates of $\Re^4$ refer to the $(x,y)$ coordinates of the curves in $\Re^2$, while the other two coordinates refer to the primary parameters $(a,b)$ of $f_S$. At the beginning, this specific subset of $\Re^4$ is assumed to be the Cartesian product of the $(x,y)$-space with the $(a,b)$-space. But, in the following, given that we need to calculate distances between OP and all possible deformed versions of S, we treat the space of the level sets of $f_S$ as a differentiable manifold. Via the curvature of the tangent space of this manifold, we look for the curve $S$ which optimally fits OP along the normal vectors $\vec n_S$ as it is described in Sects, \ref{sec:5_1}, \ref{sec:5}. The aforementioned approach and the resulting criterion form an efficient procedure that simultaneously offers the primary parameters of the prototype curve and its optimal relative position with OP.\\
\textsl{Step 5 - Determining the different types of potential stencil-guides}\\
After application of the fitting process outlined in Step 4, each object part corresponds to a prototype curve part of a specific implicit functional form $f_S$ with primary parameters $(a,b)$. Then, we apply standard clustering techniques in order to group the estimated primary parameters of the same class of prototype curves into clusters of neighboring parameters values. We assume that each such cluster corresponds to a specific stencil-guide. In other words, we assume that all object parts that correspond to the same cluster of primary parameters have been generated by one stencil - guide, the optimal cluster representative.\\
\textsl{Step 6 - Determination of the optimal cluster representative, i.e. of the potential guides parameters}\\
Consider a cluster $Q_S^C$ and the set of object parts that correspond to it, $OP_S^C$. Then, we redefine the sub-domain between the level sets of the prototype curves $S$ and the image of $OP_S^C$ on the $f_S$-manifold, as it is determined via the estimated $(a,b) \in Q_S^C$. Application of the resulting criterion, (see Sect. \ref{sec:7}), offers a unique prototype curve with specific primary parameters, to which all $OP_S^C$ are optimally fitted.

We stress that the aforementioned procedure accounts for all possible prototype curve versions, described by the functional form $f_S$ and acts on the manifold they define independently of rotation and translation transforms. So, the fitting process is independent of the initial orientation of the object parts on the $xy$-plane. Moreover, by computing the fitting error as an integral on the manifold of a class of prototype curves and by performing the error minimization exhaustively, the developed procedure offers very consistent fitting results, even in the case that the considered drawings suffer serious wear.
 \begin{figure*}[!t]
\centerline{
  \includegraphics[width=0.75\textwidth]{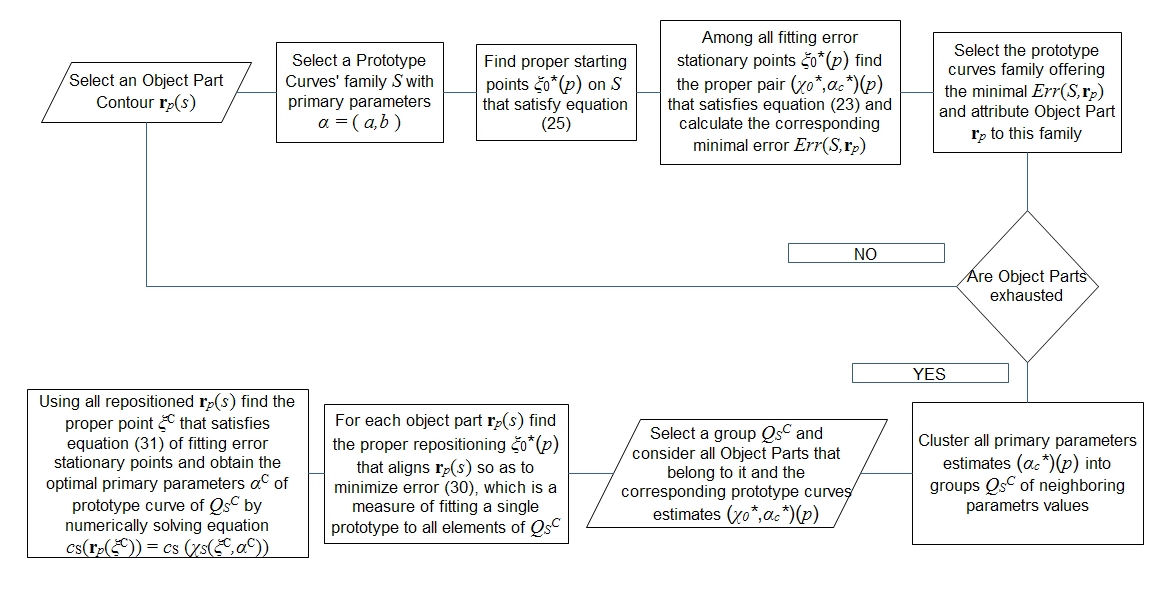}}
\caption{Block Diagram outline of the developed methodology}
\label{fig:Outline}       
\end{figure*}
 
\section{Differentiable Manifold Analysis for Fitting the Outlines of a Painting to Prototype Curves}
\label{sec:3}
\subsection{Determination of the most probable stencil part that created an object part}
\label{sec:4}
Let an object part consist of $N_p$ pixels with point vectors $\vec r _p (i)$, $i=1...N_p$. Then the contour length $s_i$ that corresponds to each point vector $\vec r_p(i)$ is approximated via $s_i = \sum _{k=1}^{i-1} {\left\| \vec r_p(k+1) - \vec r_p(k) \right\|}$. So, we can express the object part contour with curve parameter the length $s$, namely $\vec r_p(s_i)$.

	If the painter had used a part of a specific prototype curve to draw this object part, we expect that the drawn contour and the corresponding part of the prototype curve should have close curvature values along them. So let a prototype curve (stencil) $S$ given by the equation $f_S(x,y) = 0$. Curvature at length $s$ of this stencil $c_S(s) = \ddot x_S(s) \dot y_S(s) - \ddot y_S(s) \dot x_S(s)$, is also obtained via the relation \\ $c(x_S(s),y_S(s)) = \left. \nabla \cdot \left(\frac{\nabla f_S}{\left\| \nabla f_S \right\|} \right)\right|_{(x_S(s),y_S(s))}$.
	
	In order to compute the minimum distance of an object part from a properly prototype curve we adopt the flat version of the curvature $c(x,y) = \left. \nabla \cdot \left(\frac{\nabla f_S}{\left\| \nabla f_S \right\|} \right)\right|_{(x,y)}$ and for any point $(x_p , y_p)$ of the object part we compute the value $c(x_p,y_p)$. Assume, for a moment, that we properly place the object part around the adopted prototype curve so as to minimize the distances $\vec \delta (s) = \delta (s) \left(\frac{\nabla f_S}{\left\| \nabla f_S \right\|} \right)_{(x_S(s),y_S(s))}$ between them. For the computation of the integral of these minimum distances along the prototype curve we define the domain $\Omega$ between the prototype curve and the object part. Then, using Stokes' Theorem \cite{Stokes} for the integral of $c(x,y)$ on $\Omega$, we obtain 
\begin{equation}
\begin{array}{ll}
	I_1(s_0,S,p) = \int\int _{\Omega} {|c(x,y)| d \Omega}=
	\\
	= \int\int _{\Omega} {\nabla \cdot \left(\frac{\nabla f_S}{\left\| \nabla f_S \right\|} \right) d \Omega} = \oint _{\partial \Omega} {\left| \frac{\nabla f_S}{\left\| \nabla f_S \right\|} \cdot \vec \delta (s) \right| ds}
\end{array}
\end{equation}
\begin{equation}
I_1(s_0,S,p) = 2 \int _{s_0}^{s_0+L_p} {\delta (s) ds}
\label{eq:2}
\end{equation}
We also use Stokes' Theorem to obtain a boundary form for the integral of $c^2(x,y)$ on $\Omega$, as described below:
\begin{equation}
\begin{array}{ll}
I_2(s_0,S,p) = \int\int _{\Omega} {c(x,y)^2 d \Omega} = \\ = \int\int _{\Omega} {\nabla \cdot \left(c \frac{\nabla f_S}{\left\| \nabla f_S \right\|} \right) d \Omega} - \int\int _{\Omega} {\nabla c ^T \frac{\nabla f_S}{\left\| \nabla f_S \right\|} d \Omega}
\end{array}
\label{eq:3}
\end{equation}
Since $\hat n = \frac{\nabla f_S}{\left\| \nabla f_S \right\|}$ is the unit normal vector of $f_S(x,y)$ isocontours, the area differential of $\Omega$ can be written $d \Omega = d \delta \hat n ds \hat l$, where $\hat l \bot \hat n$ and $\left\| \hat l \right\| = 1$. So,
\begin{equation}
\int\int _{\Omega} {\nabla c(x,y) ^T \frac{\nabla f_S}{\left\| \nabla f_S \right\|} d \Omega} = \oint _{\partial \Omega} {\left[c(x,y) \right]_{(x_S,y_S)}^{(x_p,y_p)}ds} = 0
\label{eq:4}
\end{equation}
Then, by applying Stokes' Theorem in (\ref{eq:3}) we obtain
\begin{equation}
I_2(s_0,S,p) = \int _{s_0}^{s_0+L_p} {\left| c(\vec r_p(s-s_0)) - c(\vec r_S(s)) \right| ds}
\label{eq:5}
\end{equation}

In order to obtain the most possible part of stencil $S$ used for drawing the considered object part, we should determine the proper $s_0$ that minimizes $I_1(s_0,S,p)$. But as it is shown in Appendix \ref{Ap1} minimization of $I_2(s_0,S,p)$ coincides with minimization of $I_1(s_0,S,p)$. Thus, the part of stencil $S$ that had, most probably, been used for drawing the considered object part consist of points $\vec r _S^* = (x_S(s),y_S(s))$, $s \in \left[s_0^* , s_0^* + L_p \right]$, where
\begin{equation}
s_0^* = \arg \min _{s_0} \sum _{i=1}^{N_p}{\left| c(\vec r_p(s_i)) - c_S(s_i+s_0) \right|}
\label{eq:5}
\end{equation}

\subsection{Definition of the primary parameters for a family of prototype curves}
\label{sec:5_0}
The analysis of previous Sect. \ref{sec:4} has offered a functional representation of the optimal fitting between an object part and a given prototype, $S$, described by the locus $\vec r \in \Re^2 : f_S(\vec r) = 0$. But, the prototype curve, $S$, can be described as a realization out of all the curves that belong to the parametrized family $f_S(x,y | \vec A)=0$; where $\vec A = [a_1,...,a_{N_A}]$ is the vector of the free parameters for this class of curves. The parameters that form vector $\vec A$ describe spatial transformations (rotation and translation) and shape (curvature) variances between the curves of the corresponding family. Since, analysis of Sect. \ref{sec:4} makes use of the flat curvature functional over the descriptive equation of the prototype curve, we should determine the number of parameters which are sufficient to describe all curvature variances between the curves of the same family.

 Namely, we should exploit the derivative $\partial_{a_n} c_S(\chi , A)$, where $\chi = (x,y)^T$, $\xi = (-y,x)^T \bot \chi$, and $c_S(\chi,A)$ is the flat curvature that corresponds to the class of curves with descriptive equation $f_S(\chi | A) = 0$. For the subsequent analysis we should recall the definition of the unit directional vectors $\hat n  (\chi) = \frac{\nabla_{\chi} f_S}{\left\| \nabla_{\chi} f_S \right\|}$ and $\hat l (\chi) = \frac{\nabla_{\xi} f_S}{\left\| \nabla_{\chi} f_S \right\|} \bot \hat n (\chi)$ employed in Sect. \ref{sec:4}, where $\nabla_{\chi}=\left(\partial_x , \partial_y \right)^T$, $\nabla_{\xi}=\left(-\partial_y , \partial_x \right)^T$. So, the flat curvature is related with $\hat l (\chi)$ and $\hat n (\chi)$ via the expressions
\begin{eqnarray}
\left(\frac{\partial}{\partial \chi} \hat n \right)^T \hat l = -c_S(\chi,A) \hat l & \left(\frac{\partial}{\partial \chi} \hat l \right)^T \hat l = c_S(\chi,A) \hat n
\label{eq:curv}
\end{eqnarray}
and the partial derivatives of the directional vectors with respect to parameter $a_n$ read
\begin{eqnarray}
\partial_{a_n} \hat n = \left(\hat l^T \frac{\nabla_{\chi} \partial_{a_n} f_S}{\left\| \nabla_{\chi} f_S \right\|}\right) \hat l & \partial_{a_n} \hat l = \left(\hat n^T \frac{\nabla_{\xi} \partial_{a_n} f_S}{\left\| \nabla_{\chi} f_S \right\|}\right) \hat n
\label{eq:curvDer}
\end{eqnarray}
Subsequently, using expression (\ref{eq:curv}) for the curvature and (\ref{eq:curvDer}) for $\partial_{a_n} \hat l$ and $\partial_{a_n} \hat n$, the $a_n$-derivative of $c_S$ reads $\partial_{a_n} c_S(\chi , A) = \partial_{a_n} \left[ \hat l^T \left(\frac{\partial}{\partial \chi} \hat n \right) \hat l \right]= \hat l^T \left(\frac{\partial}{\partial \chi} \partial_{a_n} \hat n \right) \hat l$, since (\ref{eq:curv}) and (\ref{eq:curvDer}) imply that $\hat l^T \left(\frac{\partial}{\partial \chi} \hat n \right)\partial_{a_n} \hat l = \left(c_S \hat l \right)^T$ \\ $\left(\partial_{a_n} \hat l \right) = 0$. Using equation (\ref{eq:curvDer}) for $\partial_{a_n} \hat n$ and next (\ref{eq:curv}) for $\hat l^T \frac{\partial}{\partial \chi} \hat l$ we obtain
\begin{equation}
\partial_{a_n} c_S = \frac{\hat l^T \left(\frac{\partial^2}{\partial \chi^2} \partial_{a_n} f_S \right) \hat l}{\left\| \nabla_{\chi} f_S \right\|} + c_S \left(\hat n^T \frac{\nabla_{\chi} \partial_{a_n} f_S}{\left\| \nabla_{\chi} f_S \right\|}\right) 
\label{eq:DaC}
\end{equation}
Next, by expanding the derivative, $\nabla_{\chi}$, of the function $f^{a_n}_S=\partial_{a_n}f_S$, one obtains $\nabla_{\chi}f^{a_n}_S=\partial_{\hat n} f^{a_n}_S \hat n + \partial_{\hat l} f^{a_n}_S \hat l$ and subsequently (\ref{eq:DaC}) results
\begin{equation}
\partial_{a_n} c_S = \frac{\hat l^T \nabla_{\chi} (\hat l^T \nabla_{\chi} f_S^{a_n})}{\left\| \nabla_{\chi} f_S \right\|}
\label{eq:DaC1}
\end{equation}
Using equation (\ref{eq:DaC1}) we obtain the differential of $c_S(\chi,\vec A)$ in the parameters space $d_{\vec A}c_S=(\partial_{\vec A}^T c_S) d \vec A=\frac{1}{\left\| \nabla_{\chi} f_S \right\|}$ $\hat l^T \nabla_{\chi} (\hat l^T \nabla_{\chi} \partial^T_{\vec A} f_S d \vec A)$, since $\hat l^T \nabla_{\chi}$ commutes with addition. But the maximal number of linearly independent vectors $\nabla_{\chi} \partial^T_{\vec A} f_S d \vec A$ is 2. Equivalently the minimal number of parameters sufficient to describe all curvature variances among the curves of the same family is 2.

In order extract the primary parameters $(a,b)$ of a curves family parametrized via a parameters vector $\vec A$ of arbitrary size, we exploit the influence of affine mapping to the flat curvature function. Namely, if $\chi=\left(\Lambda_{\chi} R\right) \chi_T + \vec d$, where $R : R^T R = I$, $\Lambda_{\chi}=\left[\begin{array}{lll} a & 0 \\ 0 & b \end{array}\right]$
\begin{equation}
c_{S}(\chi_{T},\vec A) = (ab)^2 c_{S}(\chi,\vec A) \frac{\left\| \nabla_{\chi} f_S \right\|^3}{\left\|\Lambda_{\chi} \nabla_{\chi} f_S \right\|^3}
\label{eq:Affmap0}
\end{equation}
Next, since $a$, $b$ are linearly independent, we suppose that any parameters (i.e. non-spatial) variation of $c_S$ can be obtained by variations of $(a,b)$, $d \alpha =(da,db)$. So, $d_{\vec A}c_{S}(\chi_T,\vec A) =c_{S}(\chi,\vec A)\left\| \nabla_{\chi} f_S \right\|^3 \nabla_{\alpha}^T\left(\frac{(ab)^2}{\left\|\Lambda_{\chi} \nabla_{\chi} f_S \right\|^3}\right)d \alpha$ and since $\partial_{ll}\left\| \partial_{\vec A}f_S \right\| = \partial_{ll}\partial_{\vec A}^T f_S \frac{\partial_{\vec A} f_S}{\left\| \partial_{\vec A}f_S \right\|}$
\begin{equation}
d \vec A = \hat n_{\vec A} \frac{c_{S}(\chi,\vec A)}{\partial_{ll}\left\| \partial_{\vec A}f_S \right\|} \nabla_{\alpha}^T \ln \left(\frac{(ab)^2}{\left\|\Lambda_{\chi} \nabla_{\chi} f_S \right\|^3}\right)d \alpha
\label{eq:Affmap1}
\end{equation}
where $\hat n_{\vec A}=\frac{\partial_{\vec A} f_S}{\left\| \partial_{\vec A}f_S \right\|}$ and $\partial_{ll}=\hat l^T \nabla_{\chi}(\hat l^T \nabla_{\chi})$.

Hence, given a parametrized implicit curve $f_S(\chi,\vec A)=0$, its primary parameters $\alpha = (a,b)$ can be recovered by the mapping $(x,y) \mapsto (x/a,y/b)$ and then any differential variation of the primary parameters evolves free parameters vector $\vec A$ as equation (\ref{eq:Affmap1}) describes.


\subsection{Analysis for optimally fitting an object part to a class of prototype curves}
\label{sec:5_1}
In Sects. \ref{sec:4} and \ref{sec:5_0} it has been shown that the action of the flat curvature functional over the implicit form of a class of curves $f_S(x,y,a,b)=0$ is sufficient to optimally align an object part along the curve defined by a fixed parametrization $(a,b)=\vec A$. But, still, the pair of parameters that offer the best alignment of the object part has to be determined. In order to deal with the variances of parameters $(a,b)$ we let function $f_S$ act on $\Re^4$ and on the configuration $\vec r = (x,y,a,b)$. The vectors normal to the tangent space of the manifold defined by $f_S(\vec r) =$ constant are given by $\vec n_S = \frac{\nabla_{\vec r} f_S}{\left\| \nabla_{\vec r}f_S \right\|}$.  Thus, the curvature of the manifold tangent space is given by the differential
\begin{equation}
d \vec n_S = \frac{d \nabla_{\vec r} f_S}{\left\| \nabla_{\vec r}f_S \right\|} - \frac{\vec n_S}{\left\| \nabla_{\vec r}f_S \right\|}\vec n_S^T d \nabla_{\vec r}f_S
\end{equation} 
If we project this differential on $\vec n_S$ we obtain $\vec n_S^T d \vec n_S = 0$ implying that $d \vec n_S$ lies on the tangent space of $f_S(\vec r) = $ constant. Moreover, the tangent space of $f_S(\vec r)$ can be obtained by extending the tangent spaces of the projections to $(x,y)$ or $(a,b)$ planes. Namely $\vec l_S = \frac{1}{\left\|\nabla_{\vec r}f_S \right\|} \left(\begin{array}{lll}
\partial_{\xi} + \partial_{\beta}, & \partial_{\chi^{\alpha}}-\partial_{\alpha^{\chi}},
& -\partial_{\xi^{\beta}} + \partial_{\beta^{\xi}} \end{array}\right) f_S =$\\ $(\vec l_{\chi} + \vec l_{\alpha}, \vec l_{\chi^{\alpha}} - \vec l_{\alpha^{\chi}}, -\vec l_{\xi^{\beta}}+\vec l_{\beta^{\xi}}) = (\vec l_{(\chi,\alpha)},\vec l_{[\chi,\alpha]},\vec l_{[\beta,\xi]})$, where $\chi^{\alpha}=$$(0,0,x,y)^T$ , $\chi=(x,y,0,0)^T$, $\xi=(y,-x,0,0)^T$, $\xi^{\beta}=(0,0,y,-x)^T$ $\alpha^{\chi}=(a,b,0,0)^T$, $\alpha=(0,0,a,b)$, $\beta=(0,0,-b,a)^T$, $\beta^{\xi}=(-b,a,0,0)$. Hence, the curvature of the tangent space is expressed via the differential vector $d \vec n_S = \vec l_S^T \frac{\partial \vec n_s}{\partial \vec r} \left(d \gamma , d r_{\chi}^{\alpha} , d \gamma_{\xi}^{\beta} \right)^T$ with flat representation
\begin{eqnarray}
\text{Tr} \left(d \vec n_S \right) = \left( \vec l_{(\chi,\alpha)}^T \frac{\partial \vec n_s}{\partial \vec r}\vec l_{(\chi,\alpha)} d \gamma + 
\vec l_{[\chi,\alpha]}^T \frac{\partial \vec n_s}{\partial \vec r}\vec l_{[\chi,\alpha]} d r_{\chi}^{\alpha} \right. \nonumber \\ \left. 
+ \vec l_{[\xi,\beta]}^T \frac{\partial \vec n_s}{\partial \vec r}\vec l_{[\xi,\beta]} d \gamma_{\xi}^{\beta} \right)
\label{eq:dnF}
\end{eqnarray}
where $\gamma = \xi + \beta$, $r_{\chi}^{\alpha}= \chi^{\alpha} - \alpha^{\chi}$, $\gamma_{\xi}^{\beta} = - \xi^{\beta} + \beta^{\xi}$.

But the differential vectors $d \chi^{\alpha}$ and $d \alpha^{\chi}$ are related with $d \alpha$ and $d \chi$ via their covariance with respect to function $f_S(\chi,\alpha)$
\begin{equation}
d \chi^{\alpha} = \frac{\partial^T \chi^{f_S}}{\partial \alpha} d \alpha \; , \; d \alpha^{\chi} = \frac{\partial^T \alpha^{f_S}}{\partial \chi} d \chi\
\label{eq:covD}
\end{equation}
where $d \chi^{f_S}$ and $d \alpha^{f_S}$ are the variances $d \chi$, $d \alpha$ that are normal to the isocontours of $f_S$.

Covariance of $\chi$ and $\alpha$ can be obtained by the demand that the total differential satisfies $df_S = \partial^T_{\chi}f_S d \chi + \partial^T_{\alpha}f_S d \alpha = 0$, which gives
\begin{equation}
\frac{\partial}{\partial \alpha}\chi^{f_S}= -\frac{\partial_{\chi} f_S \partial^T_{\alpha} f_S}{\left\| \partial_{\chi} f_S \right\|^2} \; ,\;
\frac{\partial}{\partial \chi}\alpha^{f_S}= -\frac{\partial_{\alpha} f_S \partial^T_{\chi} f_S}{\left\| \partial_{\alpha} f_S \right\|^2}
\label{eq:covF}
\end{equation}
By substituting (\ref{eq:covF}) into (\ref{eq:covD}) we obtain
\begin{equation}
d \chi^{\alpha}=-\frac{\partial_{\alpha} f_S \partial^T_{\chi} f_S}{\left\| \partial_{\chi} f_S \right\|^2} d \alpha \; , \; 
d \alpha^{\chi}=-\frac{\partial_{\chi} f_S \partial^T_{\alpha} f_S}{\left\| \partial_{\alpha} f_S \right\|^2}d \chi
\label{eq:covDF}
\end{equation}

Equivalently, $d \xi^{\beta}$ and $d \beta^{\xi}$ are related with $d \beta$ and $d \xi$ by an analogous concept of covariance between $d \xi$ and $d \beta$ with respect to function $f_S$, thus obtaining
\begin{equation}
d \xi^{\beta}=-\frac{\partial_{\beta} f_S \partial^T_{\xi} f_S}{\left\| \partial_{\chi} f_S \right\|^2} d \beta \; , \; 
d \beta^{\xi}=-\frac{\partial_{\xi} f_S \partial^T_{\beta} f_S}{\left\| \partial_{\alpha} f_S \right\|^2}d \xi
\label{eq:covDF2}
\end{equation}

Having determined all quantities related with $d \vec n_S$, we can return to (\ref{eq:dnF}) and reformulate it as a differential form using basis $d \vec \Gamma = (d \eta, d \gamma, d r_{\chi}^{\alpha}, d \gamma_{\xi}^{\beta})=d \eta \vec n_S + d \gamma \vec l_{(\chi,\alpha)} + d r_{\chi}^{\alpha} \vec l_{[\chi,\alpha]} + d \gamma_{\xi}^{\beta} \vec l_{[\xi,\beta]}$, which defines the volume form $d \Omega = d \eta \wedge d \gamma \wedge d r_{\chi}^{\alpha} \wedge d \gamma_{\xi}^{\beta}$, the line form $d \Gamma = d \eta + d \gamma + d r_{\chi}^{\alpha} + d \gamma_{\xi}^{\beta}$ and its Hodge-star dual area form $dS = \ast d \Gamma = d \gamma \wedge d r_{\chi}^{\alpha} \wedge d \gamma_{\xi}^{\beta} - d \eta \wedge d r_{\chi}^{\alpha} \wedge d \gamma_{\xi}^{\beta} + d \eta \wedge d \gamma  \wedge d \gamma_{\xi}^{\beta} - d \eta \wedge d \gamma \wedge d r_{\chi}^{\alpha}$.
Next, we define the flat curvature of the tangent space as
\begin{equation}
\kappa_S = \kappa_{\gamma} + \kappa_{r^{\alpha}_{\chi}} + \kappa_{\gamma^{\beta}_{\xi}}
\end{equation}
where $\kappa_{\gamma} = \vec l_{(\chi,\alpha)}^T \frac{\partial \vec n_s}{\partial \vec r}\vec l_{(\chi,\alpha)}$, $\kappa_{r^{\alpha}_{\chi}} = \vec l_{[\chi,\alpha]}^T \frac{\partial \vec n_s}{\partial \vec r}\vec l_{[\chi,\alpha]}$, $\kappa_{\gamma^{\beta}_{\xi}}=\vec l_{[\xi,\beta]}^T \frac{\partial \vec n_s}{\partial \vec r}\vec l_{[\xi,\beta]}$.
 Then $\kappa_S d \Omega = d \vec n_s \wedge \ast d \Gamma  =  \vec d(* \eta \vec n_S^T d \vec \Gamma)= \vec d(\eta d \gamma \wedge d r_{\chi}^{\alpha} \wedge d \gamma_{\xi}^{\beta})$, where $\vec d$ denotes the exterior derivative. Consequently, the integral of the flat curvature $\kappa_S$ over a domain $\Omega$, by exploitation of the Stokes theorem, reads
 \begin{equation}
 \int_{\Omega}\kappa_S d \Omega = \int_{\partial \Omega}\eta(\gamma,r_{\chi}^{\alpha},\gamma_{\xi}^{\beta})d \gamma \wedge d r_{\chi}^{\alpha} \wedge d \gamma_{\xi}^{\beta}
\label{eq:int_curv} 
 \end{equation}
Namely, the integral of the flat curvature over a sub-domain $\Omega$ in $\Re^4$ is equal to the integral of the distances between points of its boundary $\partial \Omega$ along the normals of $f_S$. This property will be used later in order to determine the fitting error between a family of prototype curves and the image of an object part in the curves' family manifold.

In order to avoid integration of the flat curvature over the whole domain $\Omega$, we also adopt the integral of $K^2_S d \Omega = d \vec n_S \wedge * d \vec n_S = \vec d (* K_S \vec n_S^T  d \vec \Gamma)$, where evidently $K_S =\sqrt{ \kappa_{\gamma}^2 + \kappa_{r^{\alpha}_{\chi}}^2 + \kappa_{\gamma^{\beta}_{\xi}}^2}$. Consequently application of the Stokes theorem to the integral of $K^2_S$ over a domain $\Omega$ gives
 \begin{equation}
 \int_{\Omega}K_S^2 d \Omega = \int_{\partial \Omega} \sqrt{ \kappa_{\gamma}^2 + \kappa_{r^{\alpha}_{\chi}}^2 + \kappa_{\gamma^{\beta}_{\xi}}^2} d \gamma \wedge d r_{\chi}^{\alpha} \wedge d \gamma_{\xi}^{\beta}
\label{eq:int_curv2} 
 \end{equation}

Consider now the domain $\Omega \subset \Re^4$ so as its boundary is determined by the following three elements: 
\begin{enumerate}
	\item the interpolated data points' contour $\vec r_p(\xi,a,b)=(x_p(\xi),y_p(\xi),$ $a,b)$ as it was obtained in Sect. \ref{sec:4},
	\item a segment $\vec r_S(\xi,a,b)$ $=(x_S(\xi),y_S(\xi),a,b)$ of a prototype curve,
	\item a strip along the curve of parameters $\alpha(x,y,\delta_{\alpha},\beta) = \alpha_S(\beta,\chi ) + \delta_{\alpha}\frac{\partial_{\alpha} f_S}{\left\| \partial_{\alpha} f_S \right\|}$, $\delta_{\alpha} \in [-M_{\alpha}(\beta) , M_{\alpha}(\beta)]$, $ \beta \in [\beta_0 , \beta_0 + L_{\alpha}]$, where $\alpha_S(\beta,\chi)\in \left\{\alpha : f_S(\chi,\alpha) = 0 \right\}$. 
\end{enumerate}
Then, at a point $\vec r_S(\xi)$ of the prototype curves sub-domain, let the minimal distance of the aligned data points sub-domain be $\vec \delta(\xi,a,b) = \eta(\xi,a,b)$ $\vec n_S(\xi,a,b)$. The integral of these distances over the prototype curves family sub-domain $\partial \Omega_S$ can be obtained using (\ref{eq:int_curv}) and by means of the following formula
\begin{eqnarray}
E_1(\partial \Omega) &=& \int_{\partial \Omega_S}\eta(\xi,a,b) d \gamma \wedge d r_{\chi}^{\alpha} \wedge d \gamma_{\xi}^{\beta} \nonumber \\ &=& \frac{1}{2} \int_{\Omega} \kappa_{\gamma} + \kappa_{r_{\chi}^{\alpha}} + \kappa_{\gamma_{\xi}^{\beta}} d \Omega
\end{eqnarray}
The problem of finding the optimal fitting of an object part data points to a proper parametrized implicit prototype curve is now expressed via the demand of finding the $\partial \Omega$ that minimizes $E_1$. But, in Appendix \ref{Ap1} it is shown that minimization of $E_1$ demands minimization of the integral of equation (\ref{eq:int_curv2}), which we will call $E_2(\partial \Omega)$.
  These integrals consider domain $\Omega$ with respect to the manifold defined by the level-sets of $f_S(\chi,\alpha)$. So, under consideration of the basis $d \vec \Gamma$ together with the relations (\ref{eq:covDF}) and (\ref{eq:covDF2}) the volume form of $\partial \Omega$ reads 
\begin{eqnarray}
d \gamma \wedge d r_{\chi}^{\alpha} \wedge d \gamma_{\xi}^{\beta} &=&-{c_{\chi}^{\alpha}}^2 \left[\left(\frac{T_{\alpha}^2}{T_{\chi}^2}+1 \right) d \xi \wedge d \delta_{\alpha} \wedge d \beta \right. \nonumber \\ &+& \left. \left(1+ \frac{T_{\chi}^2}{T_{\alpha}^2} \right) d \delta_{\chi} \wedge d \xi \wedge d \beta \right]
\label{eq:VlFrm}
\end{eqnarray} 
where $T_{\chi} = \left\| \partial_{\chi}f_S \right\|$, $T_{\alpha} = \left\| \partial_{\alpha}f_S \right\|$, $c_{\chi}^{\alpha} = \vec n_{\chi}^T \vec n_{\alpha}$, $\vec n_{\chi} = \frac{\partial_{\chi}f_S}{T_{\chi}}$, $\vec n_{\alpha} = \frac{\partial_{\alpha}f_S}{T_{\alpha}}$, $d \delta_{\alpha} = \vec n_{\alpha}^T d \alpha$ and $d \delta_{\chi} = \vec n_{\chi}^T d \chi$.
  
  In order to re-evaluate and simplify integration on $\partial \Omega$ using basis $(d \delta_{\chi} , d \xi , d \delta_{\alpha}, d \beta)$ we should examine how integration on the level-sets of $f_S(\chi,\alpha)$ affects integration along elements of $\vec n_S$, $d \delta_{\chi}$ and $d \delta_{\alpha}$. Namely, let equation $f_S(\chi,\alpha)=f_c$ induce two curves, one on the $\chi=(x,y)$ sub-space $\chi_S(\xi, \alpha)$ and one on the $\alpha=(a,b)$ sub-space $\alpha_S(\beta, \chi)$. Then, in Appendix \ref{Ap2} it is shown that if we define the strip along the curve $\alpha_S$, $\alpha(\beta,\delta_{\alpha})$, $\beta \in [\beta_0,\beta_0+L_{\alpha}]$, $\delta_{\alpha} \in [-M_{\alpha}(\beta),M_{\alpha}(\beta)]$, so as to have constant width, i.e. $M_{\alpha}(\beta)=M_{\alpha}$, $\forall \beta$, alignment errors $E1(\partial \Omega)$, $E2(\partial \Omega)$ are simultaneously minimized by a pair $(\xi_0^*,\beta_0^*)$ that satisfies
\begin{equation}
\int_{\beta_0^*}^{\beta_0^*+L_{\alpha}} \left. c_{\chi}^{\alpha}\frac{T_{\chi}}{T_{\alpha}}\right|_{\xi_c,\beta_c} \int_{\xi_0}^{\xi_0+L_p}\left| \varepsilon_{\chi}(\xi,\beta_c) \right|  d \xi d \beta_c = 0
 \label{eq:DLang}
\end{equation}
for some $\xi_c \in [\xi_0^*,\xi_0^*+L_p]$, and
\begin{equation}
(\xi_0^{*},\beta_0^{*}) = \arg \underset{(\xi_0,\beta_0)} \min \int_{\beta_0}^{\beta_0+L_{\alpha}} \int_{\xi_0}^{\xi_0+L_p}\left| \varepsilon_{\chi}(\xi,\beta_c) \right|  d \xi d \beta_c
\label{eq:MinLang}
\end{equation}
where $\varepsilon_{\chi}$ is defined via the formula
\begin{equation}
\varepsilon_{\chi}(\xi,\beta_c)=\left[K_S {c_{\chi}^{\alpha}}^2 \left(1+\frac{T_{\alpha}^2}{T_{\chi}^2}\right)\right]_{\vec r_S(\xi,\alpha_S(\beta_c))}^{\vec r_p(\xi-\xi_0,\alpha_S(\beta_c))}
\label{eq:epsilonN}
\end{equation}
  
 \begin{figure}[!t]
  \includegraphics[width=0.45\textwidth]{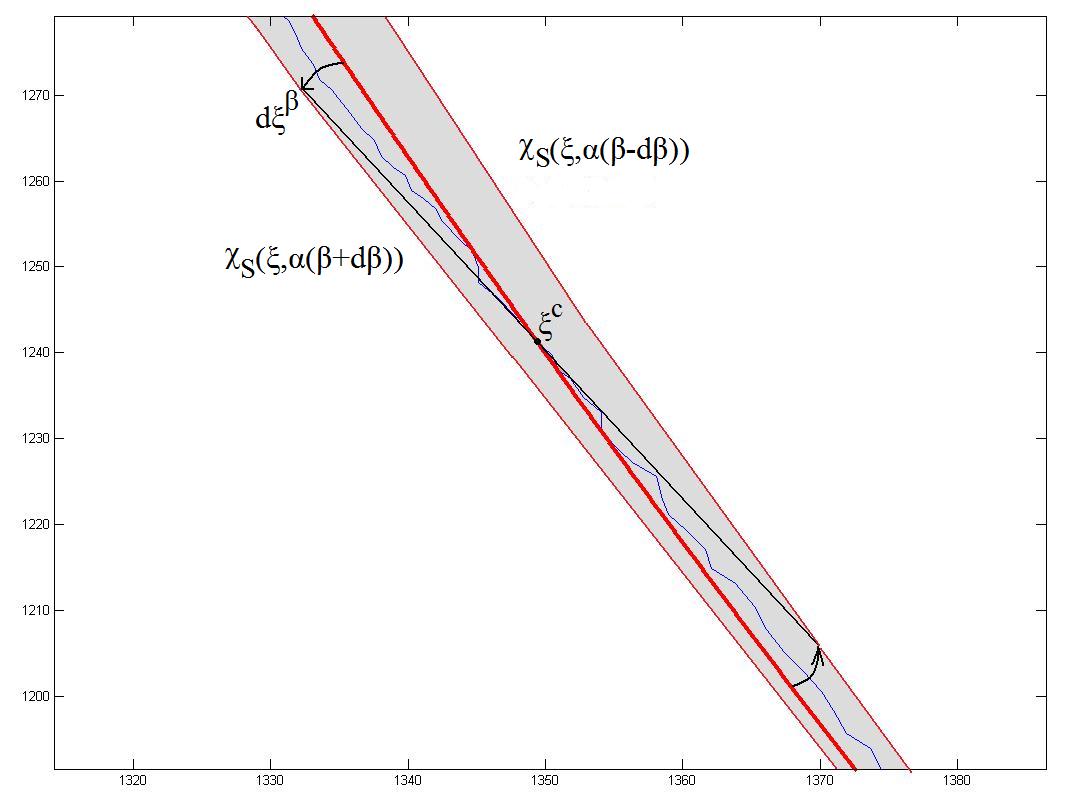}
\caption{Differential domain between an object part and a family of prototype curves. The fitting process starts by letting an arbitrary prototype curve pass from a different point, $\xi_c$, of the object part each time. In the figure the considered object part is the blue line and the prototype curve is the red one. Differential movement $d \beta$ along primary parameters curve causes differential deformation of the prototype's curvature depicted with the differential arc of length $d \xi^{\beta}$. 
}
\label{fig:Omega1}       
\end{figure}

\subsection{Simultaneous determination of the primary parameters and the segment of the prototype curve that optimally fits an object part}
\label{sec:5}
In this section, the results of Sect. \ref{sec:5_1} are exploited so as to determine the optimal placement between the data points of an object part and the elements of a prototype curves family. So, let, once again, the considered object part consist of $N_p$ points $\vec r _p (\xi_i)$, $i=1,...,N_p$. 
In order to evaluate the integrals in equations (\ref{eq:DLang}), (\ref{eq:MinLang}), we, first, determine the curve differentials $d \chi_S$ and $d \alpha_S$ along the $\chi$ and $\alpha$ level sets of $f_S$ respectively. Namely, $d x_S  = \frac{\partial_y f_S}{T_{\chi}} d \xi$, $d y_S  = -\frac{\partial_x f_S}{T_{\chi}} d \xi$, $d a_S  = -\frac{\partial_b f_S}{T_{\alpha}} d \beta$, $d b_S  = \frac{\partial_a f_S}{T_{\alpha}} d \beta$,
where, as in Sect. \ref{sec:5_1}, $d \xi$ and $d \beta$ denote the curve length differentials along the $\chi$ and $\alpha$ - level-sets of $f_S$ respectively. Then, starting from the points $\alpha_0=(a_0,b_0)$, $\vec r_S(\xi_0,\alpha_0)$ and for a constant step $d \beta$ and a given parameters curve length $L_{\alpha}$ we iteratively compute quantity
$\varepsilon_{\chi}(\xi_n,\beta_m)$ using its definition in (\ref{eq:epsilonN}) with arguments $\vec r_S(\xi_{n+1},\alpha_m)= \vec r_S(\xi_n,\alpha_m)+\left. (dx_S,dy_S) \right|_{\vec r_S(\xi_n,\alpha_m),\alpha_m}$ and $\alpha_{m+1} = \alpha_m + \left. (da_S,db_S)\right|_{\vec r_S(\xi_0,\alpha_m),\alpha_m}$. Usingsing Algorithm \ref{alg:Alg1}, among all $(\xi_0,\alpha_0)$ we spot the pair $(\chi_0^*,\alpha_c^*)$ that satisfies both (\ref{eq:DLang}) and (\ref{eq:MinLang}). The fitting error between the object part and the prototype curve part determined by this procedure is calculated via $I_2(\xi_0^*,\alpha_c^*)$ as defined in Sect. \ref{sec:5_0} and by formula $Err(S,\vec r_p) = \frac{1}{L_p}\sqrt{I_2(\xi_0^*,\alpha_c^*)}$. 

	Among all the examined classes of prototype curves and the corresponding stencils obtained by the fitting procedure, we decide that each such stencil could had generated the object part in hand if the fitting error $Err(S,\vec r_p)$ is smaller than a plausible threshold, say 0.7mm/pixel. In practice, each object part matched only one of the potential prototypes with such a small error. However, we set an extra criterion for safety, namely that if an object part matches two different prototypes with such a small error, then we assume that this might have been generated by the stencil that offers the smallest matching error. In this way, when the above process has been applied to all object parts and all potential prototypes, then a one-to-one correspondence has been established between any object part and a specific potential prototype part. However, the exact value of the primary parameters of the few stencils that might had been used for drawing the wall-paintings, must yet be estimated.

%

	\begin{algorithm}
\caption{Fitting an object part to a class of prototype curves}
\label{alg:Alg1}
\begin{algorithmic}
\STATE $E_2^* \leftarrow$ maximal number 
\FORALL{$\xi_0$}
\STATE $E_2 \leftarrow \sum_{m=0}^{L_{\alpha}/d\beta}\sum_{n=0}^{N_p}\left| \varepsilon_{\chi}(\xi_n ,\beta_m) \right| d\xi_n d_\beta$
\IF{$E_2 \leq E_2^*$}
\STATE $E_2^* \leftarrow E_2$
\STATE $\xi_0^* \leftarrow \xi_0$
\ENDIF
\ENDFOR
\FORALL{$\xi_c \in [\xi_0^*,\xi_0^*+L_p]$}
\STATE $J_0 \leftarrow \sum_{m=0}^{L_{\alpha}/d\beta}\left. c_{\chi}^{\alpha} \frac{T_{\chi}}{T_{\alpha}} \right|_{\xi_c,\beta_m}\sum_{n=0}^{N_p}\left| \varepsilon_{\chi}(\xi_n ,\beta_m) \right| d\xi_n d_\beta$
\IF{$\xi_c=\xi_0$}
\STATE $J^{*} \leftarrow J_0$
\ELSIF{$J_0 J^{*}<0$}
\STATE \textbf{break}
\ENDIF
\ENDFOR
\STATE $\alpha_c^* \leftarrow \alpha_S(0,\vec r_p(\xi_c))$
\STATE $\chi_0^* \leftarrow \vec r_S(\xi_0^*,\alpha_c^*)$ 
\RETURN $(\chi_0^*,\alpha_c^*)$
\end{algorithmic}
\end{algorithm}
	
 \begin{figure*}[!t]
\centerline{
\subfloat{\includegraphics[width=0.44\textwidth]{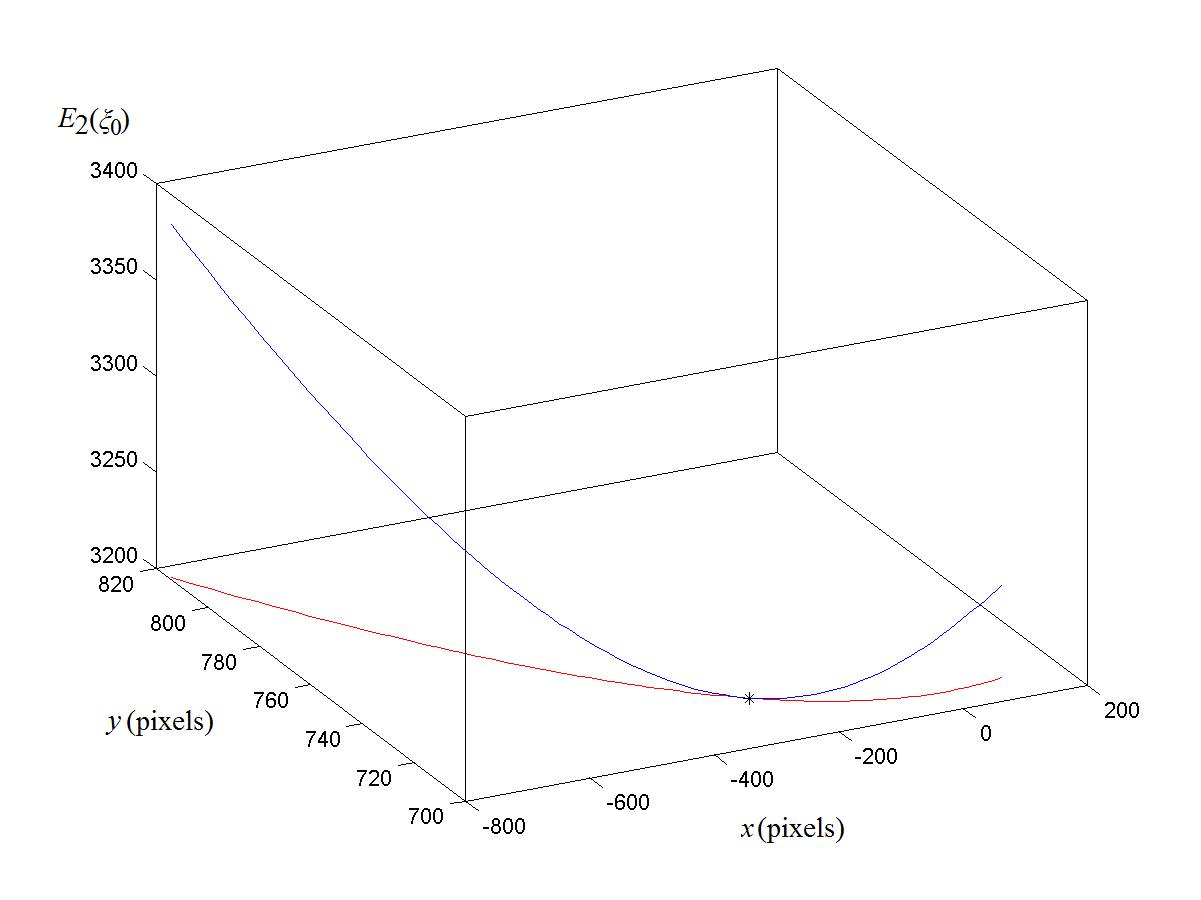} 
\label{figSubfig11}}
\hfil
\subfloat{\includegraphics[width=0.44\textwidth]{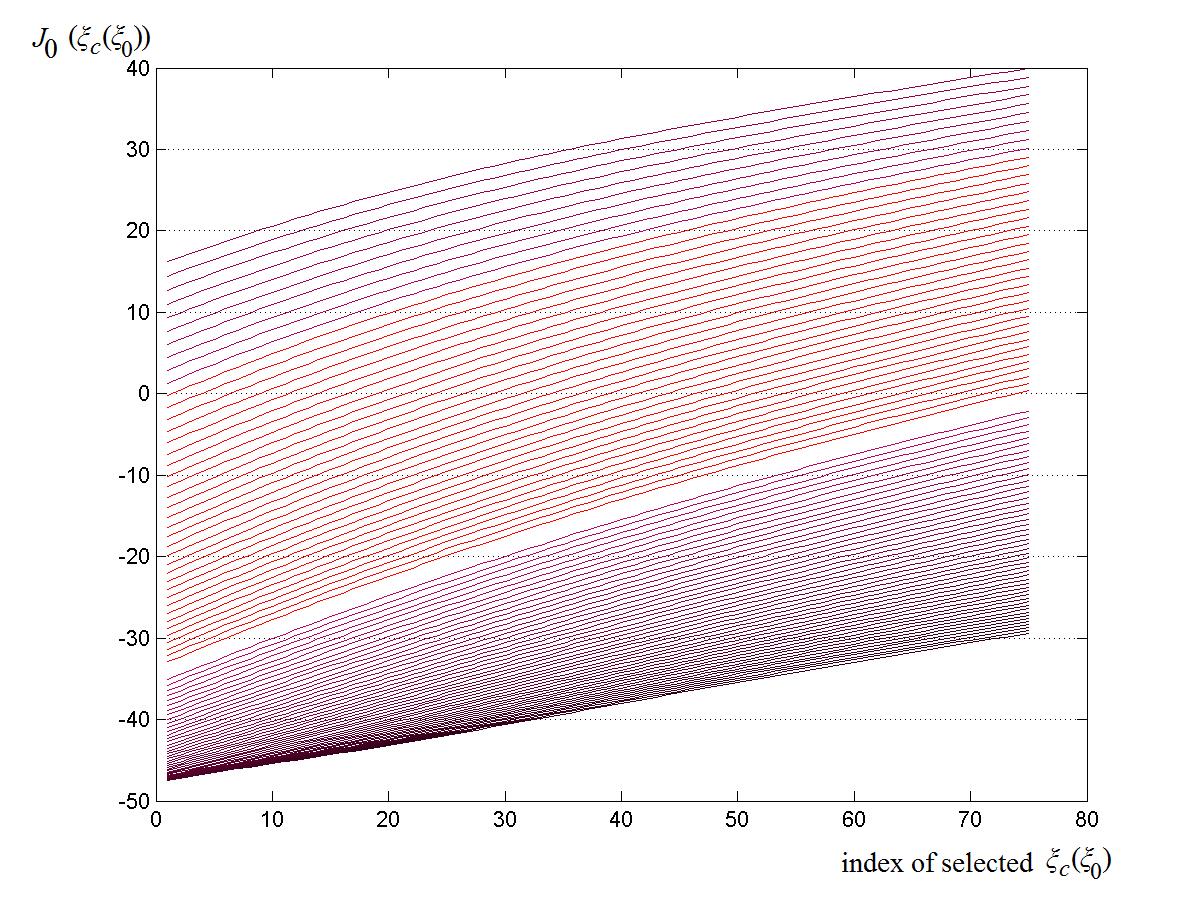}
\label{fig_subfig1_2}}}
\caption{Fitting an object part to a  prototype curve. \newline
Fig. \ref{figSubfig11} The optimal fitting error for each tested prototype curve part is depicted with the blue 3D-curve, whose $x$, $y$ coordinates correspond to the starting point 
of each prototype curve part and its $z$ coordinates correspond to the fitting error value.
The red planar curve consists of the initial points of each prototype curve part. The point 
that corresponds to the initial point of the prototype curve part that offers the minimal fitting error  for the specific object part, is depicted with an asterisk.\newline
Fig. \ref{fig_subfig1_2} Fitting error variation $J_0(\xi_c(\xi_0))$ computed via relation (\ref{eq:DLang}) for all relative placements of the considered object part and the hyperbola prototypes. Each curve of the figure corresponds to a given initial point $\xi_0$ along the prototype curves, and depicts the evaluation of $J_0$ at each point $\xi_c$ of the object part. The zero-crossings, $\xi_c^*(\xi_0)$, of these curves correspond to stationary points of the error function, $E_2$. The prototype curve parts (i.e. the $\xi_0$ selections) that offer evaluations of $J_0$ with a zero-crossing along them correspond to the curves, which are colored red.
}
\label{fig:fitting_err}       
\end{figure*}
\subsection{Clustering of the prototype curves that optimally fit the ensemble of object parts}
\label{sec:6}
Let all extracted object parts with contour point vectors $\vec r_p (s_i)$, $0 \leq s_i \leq L_p$, $p=1,...,M_{OP}$ as described in Sect. \ref{sec:2} and a subset of them $Q_S$ attributed to the family of  prototype curves denoted with $S$ via the process described in Sect. \ref{sec:5}. Moreover, for each element $\vec r_p(s_i)$ of $Q_S$, let the estimated primary parameters of the attributed prototype curve be $\alpha_p^*$. We put together all primary parameters estimations for all object parts of $Q_S$ forming a set of possible parameters vectors for the prototype $S$ over the whole wall-painting, $SA = \bigcup _{p \in Q_S} \alpha_p^*$.

In order to decide how many stencils correspond to the   family of prototype curves we classify the vectors of $SA$ via an efficient "k-Means" method \cite{kMeans}, thus obtaining $N_C$ clusters of parameters vectors $SA_C$ with minimal Euclidean distances between cluster elements. We should also emphasize that each cluster $SA_C$ of the parameters vectors has a uniquely corresponding cluster $Q_S^C$ of the object parts in $Q_S$.

Now, each cluster $SA_C$ corresponds to a potential stencil obtained from the $S$ family of prototype curves. In order to determine the parameters of these stencils and the exact part of them that probably guided the drawing of each object part we adopt the following process.

\subsection{Exact determination of the primary parameters of the prototype guides}
\label{sec:7}
In this section we will try to determine the most probable pair of primary parameters of a prototype that belongs to the class of curves $f_S(x,y,a^C,b^C) = 0$, which is supposed to have been used for drawing the object parts that belong to the cluster $Q_S^C$. In order to achieve this, we will use the analysis of Sect. \ref{sec:5_1} and Appendix \ref{Ap2} for fitting an object part to a family of prototype curves, but, this time, with the demand that $\xi_c$ and $\beta_c$, which correspond solutions of equations (\ref{eq:DLang}) and (\ref{eq:MinLang}), are kept fixed and equal to $\xi^*_p$ and $\beta^*_p$ for each object part $p$. These values correspond to the results of the fitting of each object part to the considered class of prototype curves. Thus, although $\xi_p^*$ varies between different object parts, $\beta^*_p$ is attributed to zero for obtaining any $\alpha^*_p$ (see Sect. \ref{sec:5}). Thus,we demand that the prototype curve and the object part lie on the same primary parameters level-set, with initial point $\alpha_p^*=\alpha_S(0,\vec r_p(\xi_p^*))$ and the prototype curve, $\chi_S(\xi_p,\alpha_p) = \vec r_S(\xi_p)$, estimated in Sect. \ref{sec:5_1}.

\begin{figure}[!t]
\includegraphics[width=0.45\textwidth]{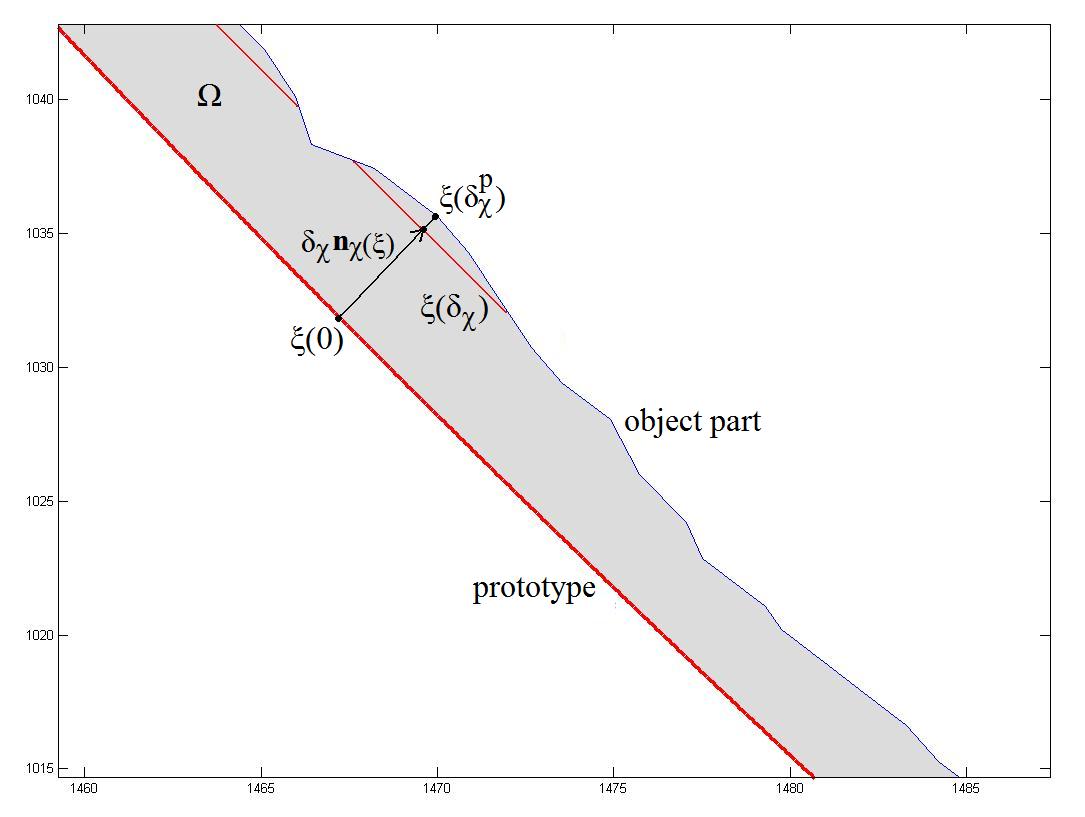}
\caption{Repositioning of an object part along a fixed prototype curve. The fitting process runs over all points of the object part (in blue) and lets them be the intersection of the object part and the prototype curve part (in red). Curve $\xi(\delta_{\chi})=\xi(\delta_{\chi}^p)$ is eventually selected. 
}
\label{fig:Omega2}       
\end{figure}

Having clustered the stencils estimated for each object part into groups of prototypes of the same kind and of neighboring primary parameters, in this section we look for the prototype curve along which all object parts, attributed to stencils of the same cluster $Q_S^C$, can be optimally fitted. Hence, the aforementioned demand for fixed primary parameters level-set for each pair of a prototype and an object part is now violated by asking for a fixed prototype representing all object parts of the same cluster. Following the analysis of Sect. \ref{sec:5_1} and the definition of domain $\Omega$ between the object part and the prototype, we will redefine the strip along the primary parameters level-sets $\alpha = \alpha_S(\beta,\chi) + \delta_{\alpha} \vec n_{\alpha}$ for given variations $\delta_{\alpha}\in [m_{\alpha},M_{\alpha}]$ along an $\alpha$ - level-set of reference and with $\beta_c(\delta_{\alpha}) \in [0,L_{\alpha}(\delta_{\alpha})]$. Consequently, we redefine the domain bounded by the prototype and an object part on the basis of the distances between the corresponding curves, $\delta_{\chi}$. Namely, each object part $p$ has been already attributed to a stencil part starting from point $\xi_0^*$ along a prototype curve with primary parameters $\alpha_p^*$ and the point $\xi^*_c(\xi_0^*)$ has been spotted, at which the prototype curve and the object part are supposed to intersect, so as the fitting error is minimized. But re-positioning the object part along a fixed prototype, does not maintain the property of $\xi^*_c$ to be an intersection of the two curves; hence, we should re-estimate $\xi_c^*$ in $[\xi_0^*,\xi_0^* + L_p]$ so as to correspond to an intersection. Moreover, the starting point $\xi_0^*$ of the prototype curve part along which the alignment error is minimized should, also, be re-estimated. Hence, for all possible $\delta_{\chi} \in [m_{\chi},M_{\chi}]$, where $m_{\chi}=\min_{\xi}\left\{(\vec r_S(\xi , \alpha) - \vec r_p(\xi))^T \vec n_{\chi}(\xi)) \right\}$, $M_{\chi} = \max_{\xi}\left\{(\vec r_S(\xi , \alpha) - \vec r_p(\xi))^T \vec n_{\chi}(\xi)) \right\}$, $\xi \in [\xi_0^*,\xi_0^* + L_p]$, we define the function $\xi(\delta_{\chi})$ which describes each possible repositioning of the object part along the prototype curve's normals so as the domain between the prototype curve and the fitted object part is covered (see Fig. \ref{fig:Omega2}).

Using this definition of domain $\Omega$, we reformulate (\ref{eq:int_curv2}) in terms of the basis $(d \delta_{\chi}, d \xi, d \delta_\alpha, d \beta)$ as in Sect. \ref{sec:5_1} to obtain the fitting error
\begin{eqnarray}
&&E_2^C(p) = \int_{m_{\alpha}}^{M_{\alpha}} \int_{0}^{L_{\alpha}(\delta_{\alpha}^C)}\int_{\xi_0}^{\xi_0+L_p} |\varepsilon_{\chi}^C| d \xi d \beta_c d \delta_{\alpha}^C \nonumber \\ 
&&+ \int_{m_{\chi}}^{M_{\chi}}\int_{\xi_0(\delta_{\chi})}^{(\xi_0+L_p)(\delta_{\chi})}\int_{\beta_0}^{\beta_0+L_{\alpha}} |\varepsilon_{\alpha}^C| d \beta d \xi_c d \delta_{\chi}^C
\label{eq:E2N}
\end{eqnarray}
where functions $\varepsilon_{\chi}^C$ and $\varepsilon_{\alpha}^C$ are given via $\varepsilon_{\chi}$ and $\varepsilon_{\alpha}$ respectively, which have been defined with equation (\ref{eq:epsilon}) of Appendix \ref{Ap2}. Boundaries $\vec r_{\xi}^{+}$, $\vec r_{\xi}^{-}$,  $\vec r_{\alpha}^{+}$, $\vec r_{\alpha}^{-}$ are determined as domain $\Omega$ defines, thus reading
$\vec r_{\xi}^{+} = \vec r_p(\xi-\xi_0)$, $\vec r_{\xi}^{-} = \vec r_S(\xi,\alpha(\beta_c,\delta_{\alpha}^C))$ and $\vec r_{\alpha}^{+} = \left(\chi(\xi_c)+\delta_{\chi}^C\vec n_{\chi} , \alpha(\beta_c,M_{\alpha}) \right)$, $\vec r_{\alpha}^{-} = \left(\chi(\xi_c)+\delta_{\chi}^C\vec n_{\chi},\alpha(\beta_c,m_{\alpha})\right)$.

In Appendix \ref{Ap3} it is shown that minimization of primary parameters estimation error (\ref{eq:E2N}) is offered by the repositioning $\xi_0^*(p)$, with intersection point at $\xi_c(\delta_{\chi}^C)$ that satisfy both to relations below
\begin{equation}
\int_{\delta_{\alpha}^C(\alpha_p^*)\in Q_S^C} \left. c_{\chi}^{\alpha}\frac{T_{\chi}}{T_{\alpha}}\right|_{\delta_{\chi}^C,\delta_{\alpha}^C} \int_{\xi_0}^{\xi_0+L_p}\left| \varepsilon_{\chi}^C(\xi,0,\delta_{\alpha}^C) \right|  d \xi d \delta_{\alpha}^C = 0
 \label{eq:DLang5}
 \end{equation}
 \begin{equation}
\xi_0^{*}(p) = \arg \underset{\xi_0} \min \int_{\delta_{\alpha}^C(\alpha_p^*)\in Q_S^C} \int_{\xi_0}^{\xi_0+L_p}\left| \varepsilon_{\chi}^C(\xi,0,\delta_{\alpha}^C) \right|  d \xi d \delta_{\alpha}^C
\label{eq:MinLang1}
\end{equation}
 
In order to evaluate the above minimizer, the points $\alpha_p^*$ are ordered along a curve from $\alpha_m^*$ to $\alpha_M^*$, where $m = \arg \min_{j} \min_{i\neq j}\left\| \alpha_i^*- \alpha_j^* \right\|$, $M = \arg \max_{j} \max_{i\neq j}$ $\left\| \alpha_i^*- \alpha_j^* \right\|$. Then point $\alpha_p^*$ is met at length $\delta_{\alpha}(p)$ of curve $d\alpha^*(\xi_0(p),\delta_{\alpha}) = \vec n_{\alpha}(\xi_0(p),\delta_{\alpha})d\delta_{\alpha}$.
The procedure that performs this minimization and returns the primary parameters pair $(a^C,b^C)=\alpha_S(0,\vec r_p^*(\xi^C))$ of the prototype that optimally fits the object parts of $Q_S^C$ is accompilshed by Algorithm \ref{alg:Alg2}.
But, this time, in order to obtain an accurate measure for the fitting error between each object part and the determined stencil part, we calculate the exact values of the point by point distances between the corresponding curves by means of the lemma introduced in \cite{PapFrag}. In fact, for each object part, we properly rotate and translate the corresponded stencil part so as their point by point distances are minimized, as the aforementioned lemma describes. For each such pair of object and stencil parts we keep record of the mean and the maximal value of their point by point distances. Such results for the considered wall-paintings are noted down in tables \ref{tab:3}, \ref{tab:4}.
\begin{algorithm}
\caption{Determination of each Object Parts group optimal prototype}
\label{alg:Alg2}
\begin{algorithmic}
\STATE $E_2^* \leftarrow$ maximal number
\FORALL{$p \in Q_S^C$}
\STATE $E_2 \leftarrow \sum_{\alpha_p^* \in Q_S^C} \sum_{n=0}^{N_p}\left| \varepsilon_{\chi}^C(\xi_n,0,\alpha^*_p)\right|  d \xi_n d \delta_{\alpha}^C(p)$
\IF{$E_2 \leq E_2^*$}
\STATE $E_2^* \leftarrow E_2$
\STATE $p^* \leftarrow p$
\ENDIF
\ENDFOR
\FORALL{$p \in Q_S^C$}
\STATE $\vec r_S(\xi^*(p),\alpha^*_p) \leftarrow \vec r_S(\xi(p) , \alpha^*_{p^*})$
\ENDFOR
\FORALL{$\xi^C \in \bigcup_{p \in Q_S^C}[\xi_0^*(p),\xi_0^*(p)+L_p]$}
\FORALL{$p \in Q_S^C$}
\STATE $J_p \leftarrow \sum_{\vec r_p \in Q_S^C}\sum_{n=0}^{N_p}\left| \varepsilon_{\chi}^C(\xi_n,0,\alpha_p^*) \right|  d \xi$
\ENDFOR
\STATE $J_0 \leftarrow \sum_{p\in Q_S^C}\left. c_{\chi}^{\alpha}\frac{T_{\chi}}{T_{\alpha}}\right|_{(\vec r_p^*(\xi^C),\alpha^*_p)} J_p d \delta_{\alpha}^C(p)$
\IF{$p=1$}
\STATE $J^* \leftarrow J_0$
\ELSIF{$J_0 J^* \leq 0$}
\STATE \textbf{break}
\ENDIF
\ENDFOR
\RETURN $\alpha_S(0,\vec r_p(\xi^C))$
\end{algorithmic}
\end{algorithm}

\begin{figure}[!t]
 \centerline{ 
   \subfloat{\includegraphics[width=0.44\textwidth]{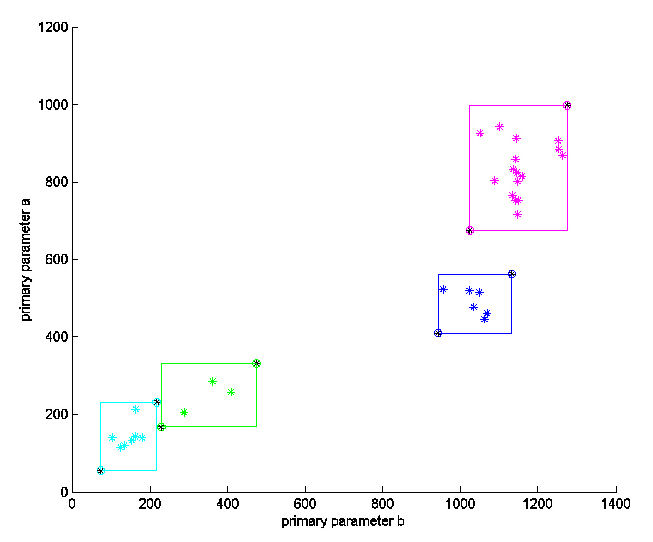}
  \label{figaPizza}}
}
\centerline{
  \subfloat{\includegraphics[width=0.23\textwidth]{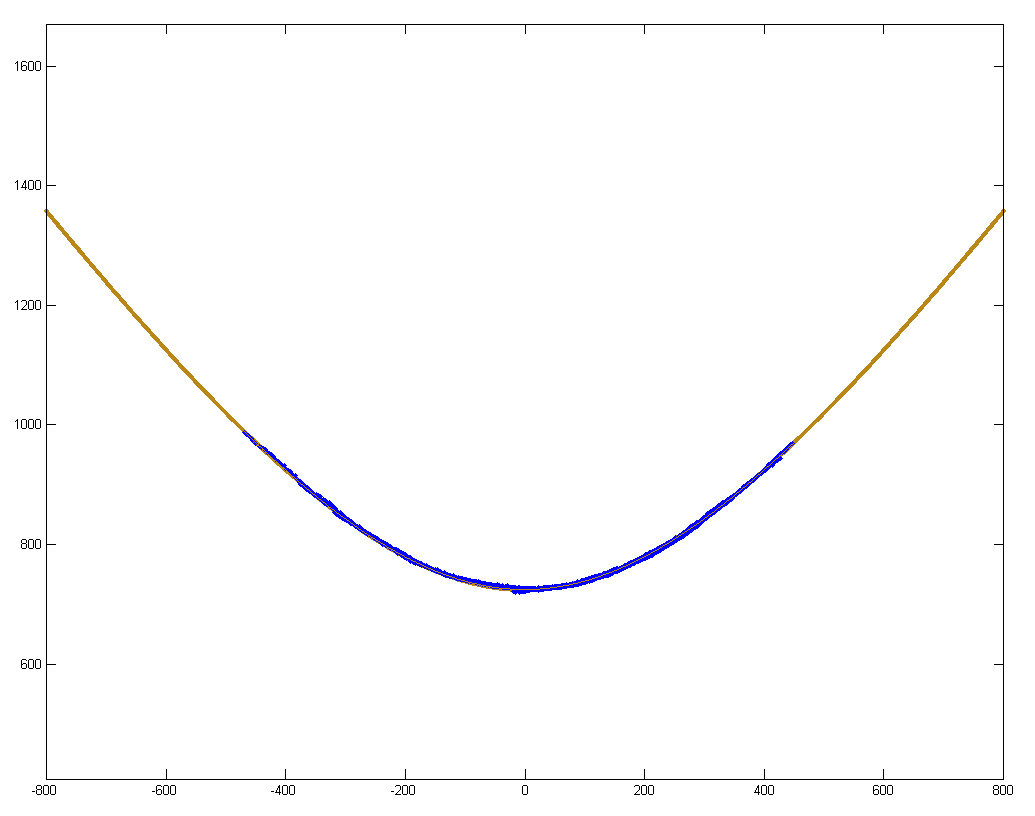} \includegraphics[width=0.23\textwidth]{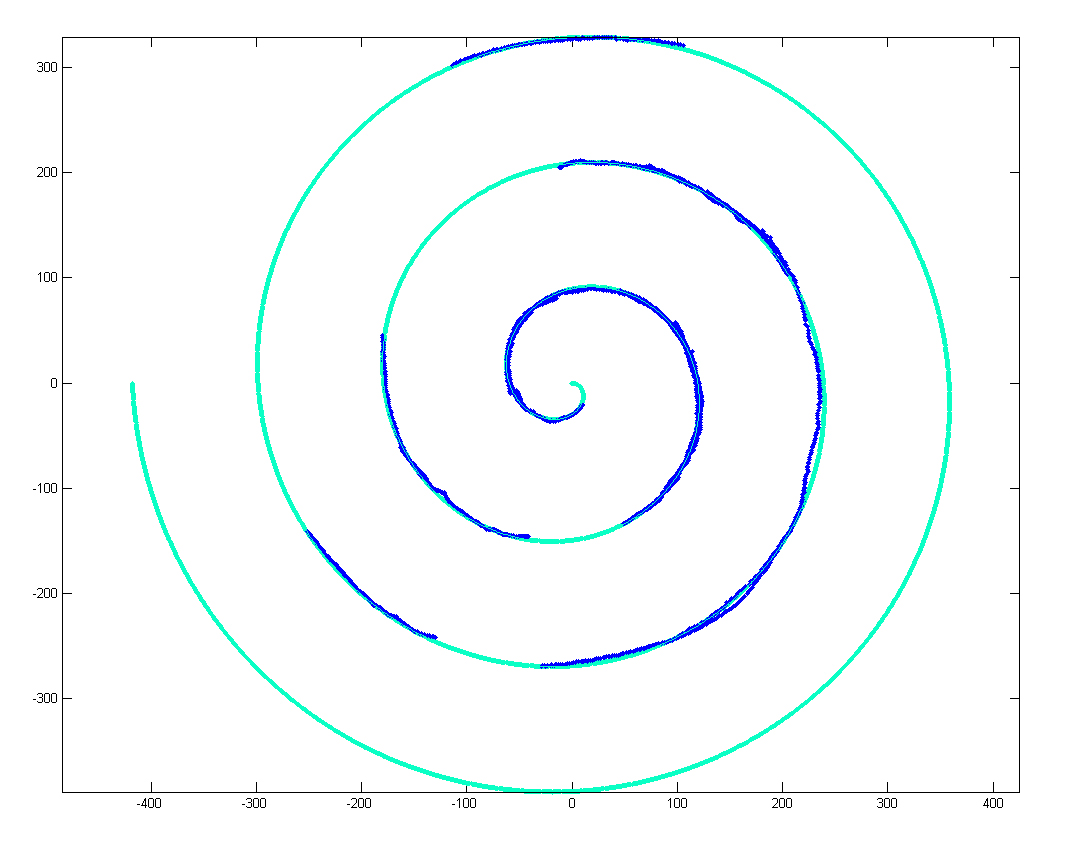}
 \label{figsten}}
  }
\caption{\ref{figaPizza}. Clustering results for the Hyperbola primary parameters determined for the Object Parts of one figure of the "Naked Boys" wall painting. The color of each group corresponds to the one attributed to its optimal guide in Table \ref{tab:2}
\newline
Fig. \ref{figsten} Optimal placement of the drawn object parts (in blue) along the prototype curves determined as optimal representatives of the corresponding object parts groups. The object parts of the figure has been extracted from the "Lady of Mycenae" wall painting and the determined prototype curves correspond to the "Hyperbola 2" and "Spiral 2" prototypes as given in Table \ref{tab:1}}
\label{fig:fit}       
\end{figure}

\section{Aspects Concerning the General Applicability of the Methodology}
\label{sec:7N}
The methodology introduced in this work has been motivated by the demands of developing a strict and quantitative test of the hypothesis for possible geometric prototypes used for drawing prehistoric wall-paintings. However, the approach and the related system deal with the problem of determining and fitting implicit curve models to contour fragments data sets in general. Namely, given a set of contour data we initially spot the contour segments of the same convexity thus forming the set of Object Parts. Then, given a set of parametrized implicit curves families these Object Parts are attributed to an exact model curve and the proper curve segments that optimally fit Object Parts are determined. As referred in the Introduction, the main advance of the curve pattern analysis and fitting methodology developed here is that a) optimal relative placement between prototype curve and data points and b) the proper form of the prototype curve that optimally fits all contour segments of its group are simultaneously offered. This is achieved by embedding all possible curvature deformations and all possible, orientation-free, relative placements between prototype curve and Object Parts into a single error functional. The introduced fitting error bares the same minima with the Euclidean one, but exploits the functional version of the curvature in order to be invariant under rotation and translation.

\begin{figure}[h]
\centerline{
  \includegraphics[width=0.4\textwidth]{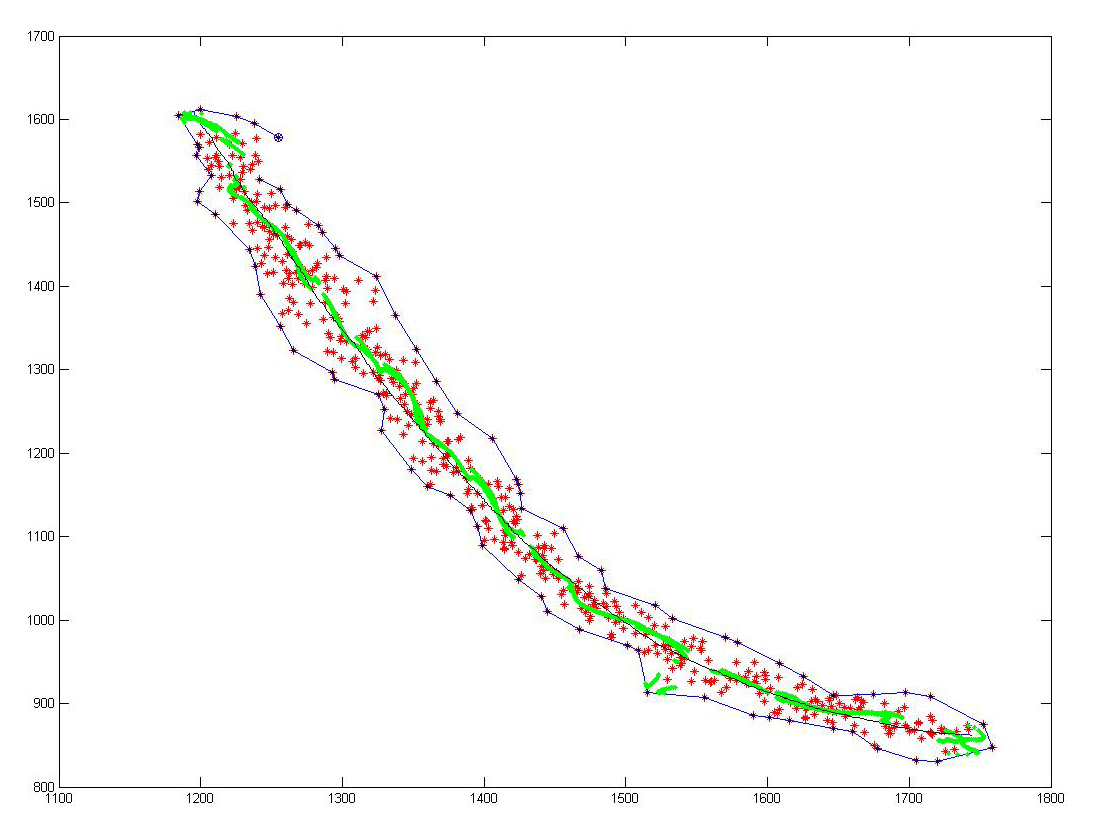}}
\caption{Construction of the envelope (blue contour) of a point cloud (red points) normally distributed around an Object Part (black contour in the center). "Mean curve" of the envelope (green points) closely approximates the original Object Part contour}
\label{fig:new0}       
\end{figure}

One immediate extension of the methodology developed in this work, concerns its application to point cloud data that do not indicate a contour form; i.e. no sense of tangent direction emerges when one reads all given data points successively. In the fitting error integrals derived by the  analysis of Sects. \ref{sec:4}, \ref{sec:4} and \ref{sec:7} the integration is performed along curve length - parametrization of the prototype curve. Hence, the contour form of Object Parts' data points is induced by the fact that for each prototype curve point there is one data point met in the direction normal to prototype curve's tangent at the specific place. So, given a set of point cloud points one could extract the contour forms used as Object Parts for the fitting process, by treating the point cloud as a compact fitting area whose boundary define the curve that should be fitted to the prototype curve. Such an approach successfully deals with normally distributed point cloud around a contour, because the mean curve of the boundary coincides with the mean curve of a normally distributed set of internal points. This case is presented in figure \ref{fig:new0}, where the contour points of an Object Part are distorted independently in $x$ and $y$ directions with Gaussian noise, thus forming a point cloud. The point cloud envelope is determined by the set of points that are assigned as the most distant ones from each point of the point cloud in all directions. Application of the fitting methodology presented here to such point clouds practically offered the original prototype curve part. 

\begin{figure}[h]
\centerline{
  \includegraphics[width=0.2\textwidth]{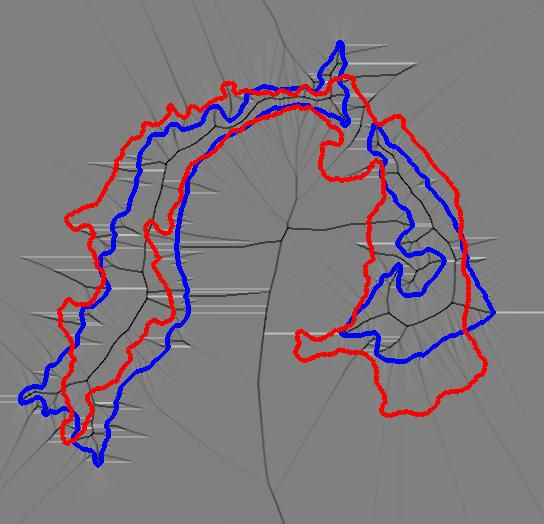}
  \includegraphics[width=0.18\textwidth]{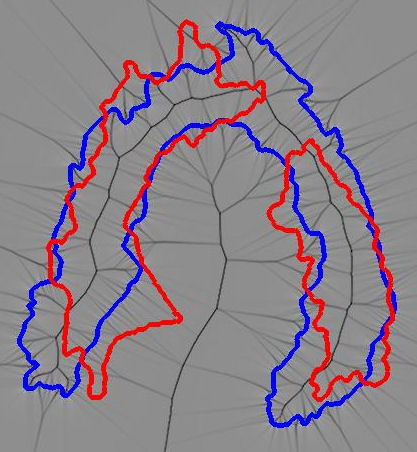}}
\caption{Evaluation of the fitting error and the resulted relative matching for two pairs of inscribed "Omega" letters. The figures present the relative matching position between prototype (blue) and data (red) contours on the space of the flat curvature evaluations of the prototype (the grayscale background). 
}
\label{fig:new1}       
\end{figure}

A second, not so immediate, extension of the methodology concerns application of the fitting error formulation and evaluation to problems of estimating similarity between shapes. In figure \ref{fig:new1} an example is presented manifesting application of the fitting methodology to estimate optimal matching between the shapes of inscribed Greek "Omega" - letter symbols. To such shape matching problems, Euclidean distance transform has been used to obtain an implicit  representation of each shape and the corresponding flat curvature values. But for a complete shape matching application there is a main issue to be resolved. Although the introduced fitting methodology offers optimal affine deformation of the curvature of a prototype shapes' family so as to match given contour data set, the fitting results evaluation relies on the selection of the prototype shapes' family. But in the case of shape matching analytic form for the prototype shape is not available and brute setting of an arbitrary shape as prototype does not necessarily offer absolutely optimal shape matching results. Hence, following optimal matching between a pair of shapes, one should join the matched shapes to a common implicit flat representation independently of the selection of which shape is considered as prototype. Such a complete extension to arbitrary shape similarity problems falls outside the goals of the present paper and it could be treated in a future work.

\begin{figure}[h]
\centerline{
  \includegraphics[width=0.19\textwidth]{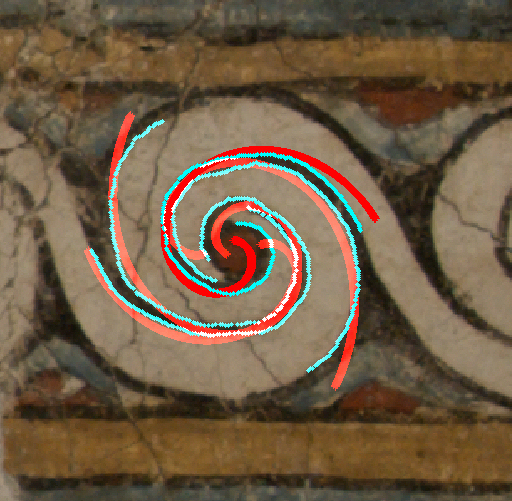}
  \includegraphics[width=0.22\textwidth]{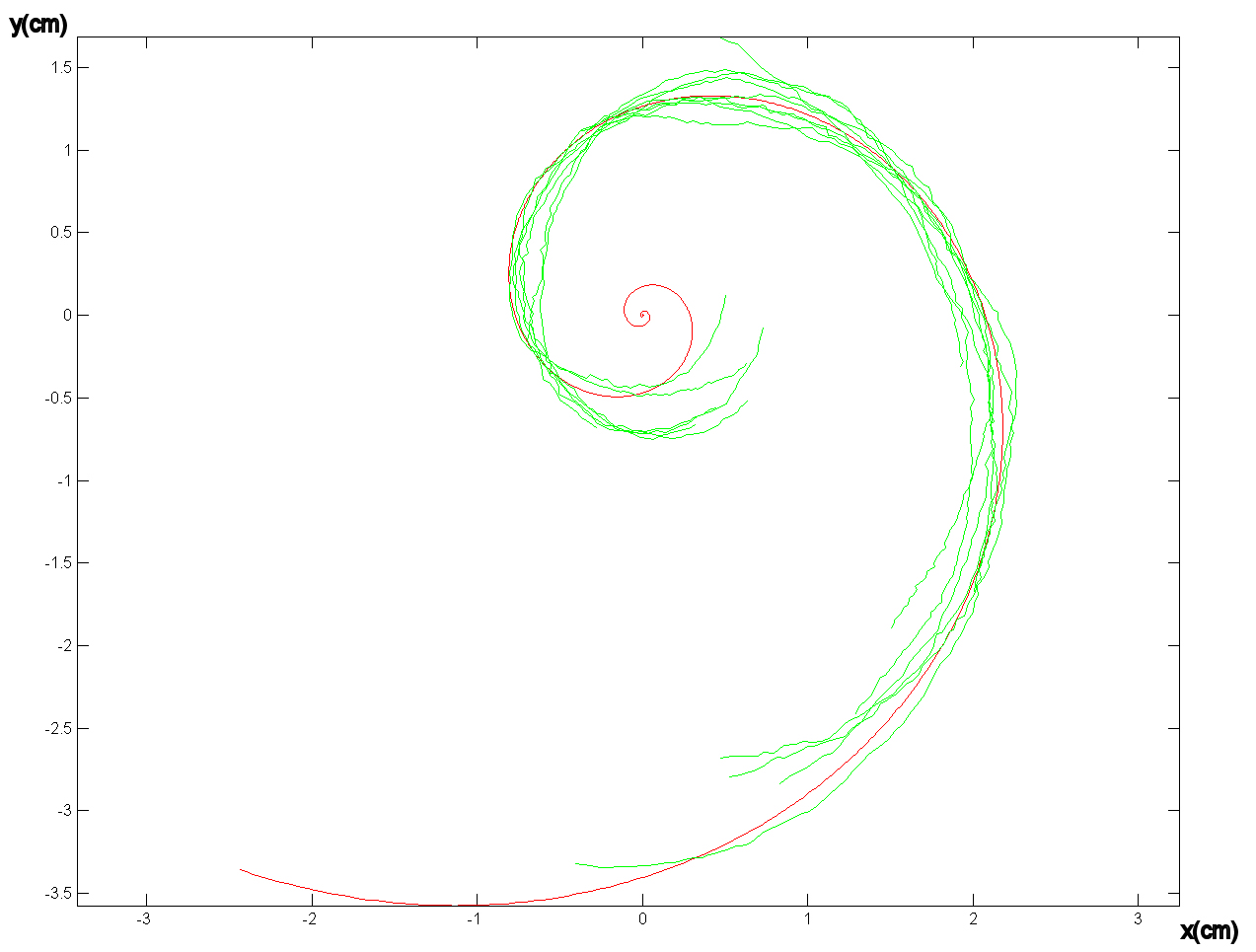}}
\caption{Rejection results for a wall-painting probably drawn by free hand. In the left figure the best fit results of drawn spirals to the considered model ones are shown. Optimal placement of the exponential spiral (in red) to the corresponding object parts (in cyan) highlights the high error values (3-6 mm/cm). In the right figure fitting of the object parts to their common prototype, depicts both optimality of the prototype determination and high error values of the fitting results. 
}
\label{fig:new1new}       
\end{figure}

Concerning the applicability of the developed methodology in cases where hypothesis for existence of prototypes in paintings, or in other application's contour data, can not be verified and thus it should be tested for rejection, we should distinguish 2 cases: a) If the implicit forms of the possible curve prototypes are given, then in case of rejection each prototype optimally fits the corresponding contour data with large fitting error. This happens because, even in the case of accidental fit between object parts and prototypes, there is no common curve that optimally fits all object parts attributed to it (see figure \ref{fig:new1new}). If such a curve exists then it is assumed to be a prototype of its object parts and the hypothesis is verified. b) If the implicit forms of the possible curve prototypes is not given, we check if the painting is made by free hand as follows: We consider all Object Parts with length greater than a proper threshold (depending on the size of the painting) and we apply to them the methodology pairwise by letting, each time, the longer Object Part of the pair to be the prototype curve. In this way we obtain a number of classes of object parts, where each class consists of the Object Parts that fit the assumed prototype with a small error. If the number of these classes is large and/or the cardinal number of each class is pretty small, then we conclude that the considered contours had been drawn by free hand.

\section{Application of the Methodology to the Determination of the Method used for Drawing Specific Prehistoric Wall-Paintings}
\label{sec:8}
\subsection{Some historical elements concerning the examined wall-paintings}
\label{sec:9_0}
	The so called Mykenaia (Lady of Mycenae), housed in the National Archaeological Museum in Athens (inv. no 11670,) is the best preserved wall painting on the Greek Mainland and of the highest quality. It comes from a Late Helladic III (13th century BC) house, near the fortification wall, in the area of the Cult Centre at Mycenae. 

The majestic face, with the almond shaped eye and the elegant profile, has a pensive expression which reveals the solemnity of the moment. It is attributed to a seated goddess who accepts the offerings of the worshippers. In this case it is a necklace made of cornelian beads, which she holds tightly in her right hand. Alternatively she could be the best preserved of a processional figure bearing offerings to a goddess. She wears a short-sleeved bodice over a sheer blouse, which delineates her ample bosom. Her intricate hairstyle and rich jewelry (necklaces and bracelets) are striking \cite{Agean}.

That a device for rendering curves was used for the curve of the nose of the Mykenaia  was noted early on \cite{ZaB}.
	\begin{figure}[!t]
\centerline{
  \includegraphics[width=0.38\textwidth]{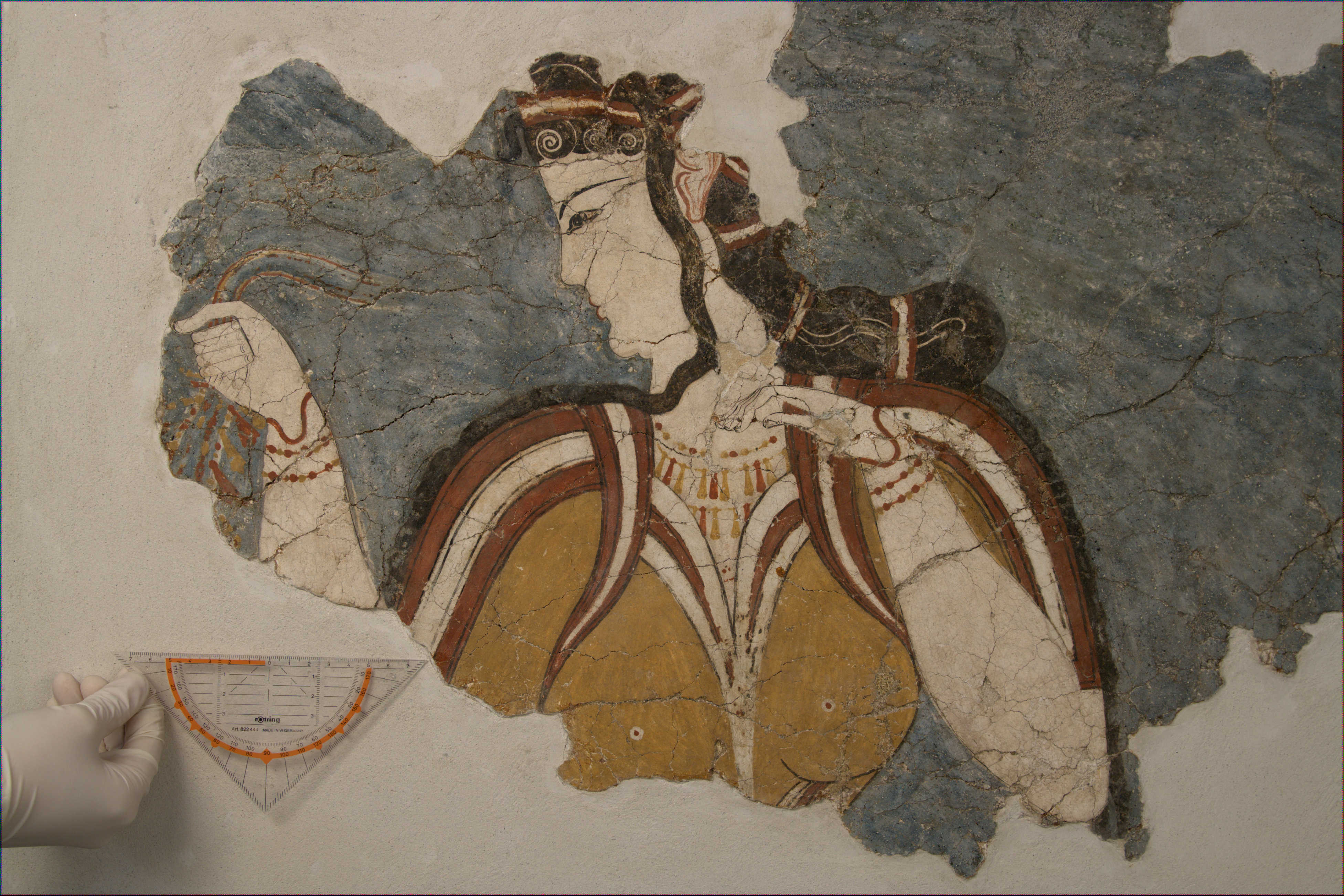}}
\caption{Photograph of the Lady of Mycenae wall-painting housed in the National Archaeological Museum of Greece.}
\label{fig:1}       
\end{figure}

	The wall-painting 'Naked Boys' \cite{Doumas} initially decorating the internal murals of the lustral basin of the edifice 'Xeste 3' excavated at Akrotiri, of the Greek island of Thera. The interpretation proposed for the scene of the wall-painting is that it depicts an initiation rite, during which at least one of the actors will achieve manhood.
\begin{figure}[h]
\centerline{
  \includegraphics[width=0.18\textwidth]{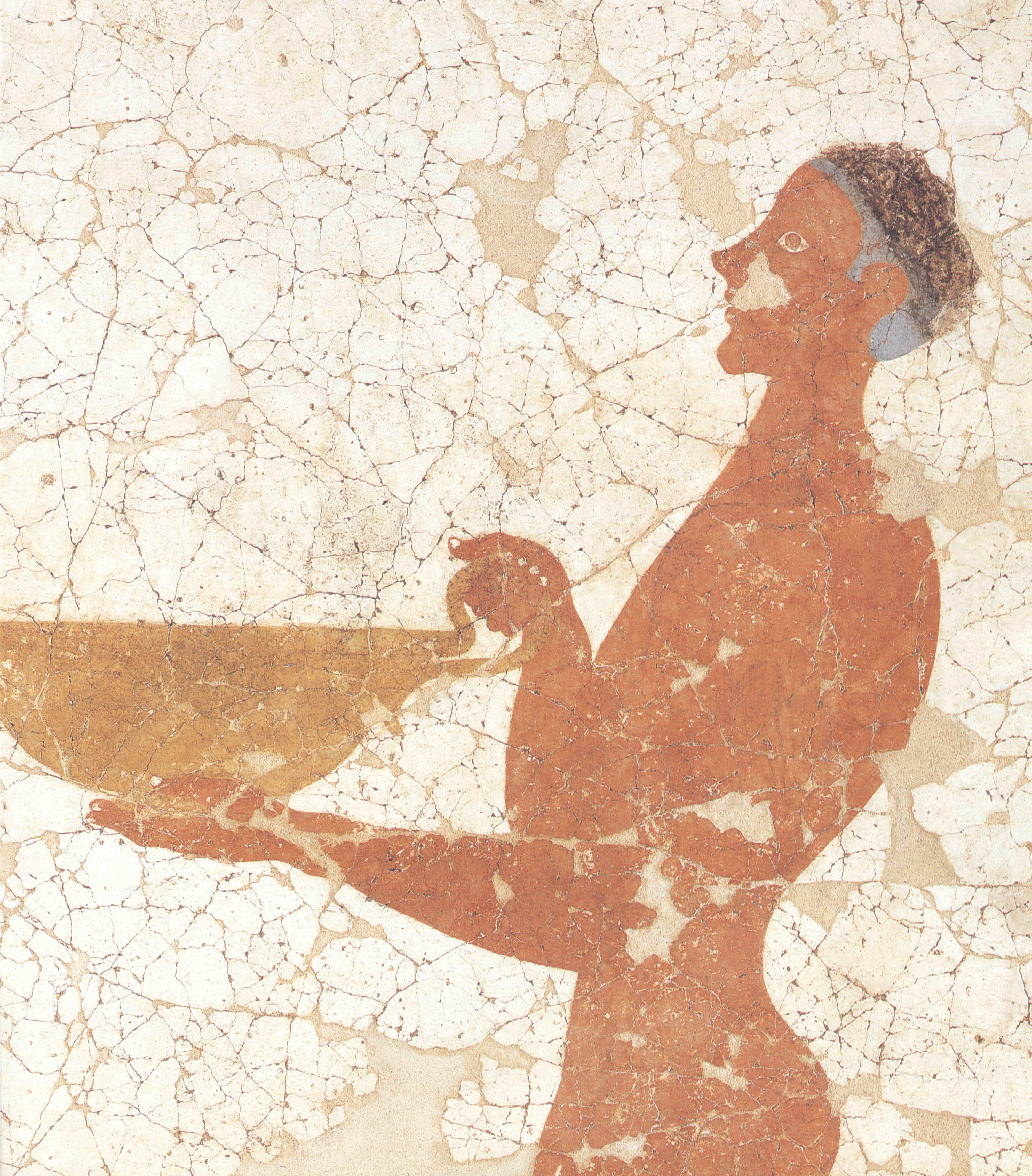}
  \includegraphics[width=0.18\textwidth]{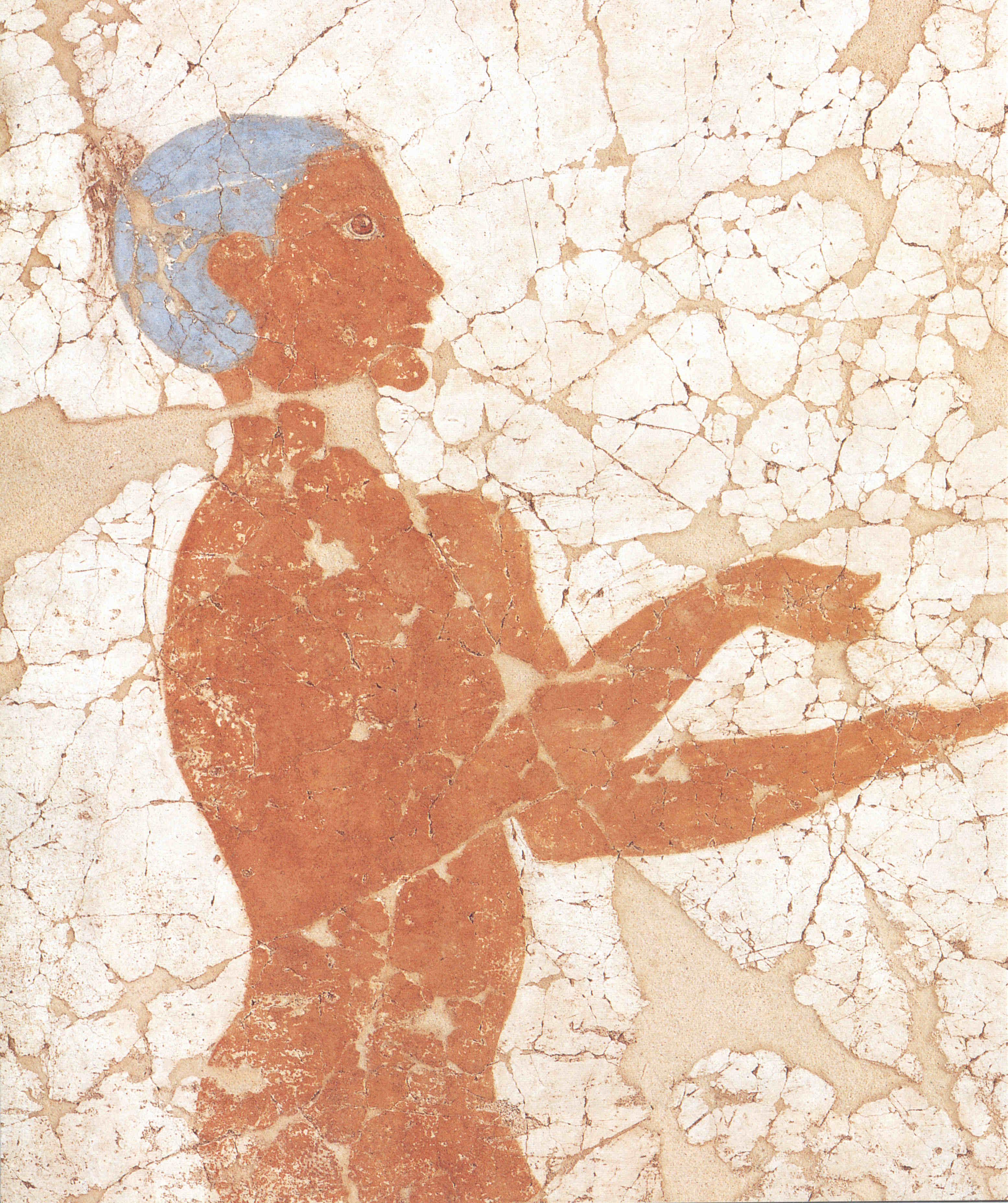}}
\caption{Photographs of two Naked Boys excavated at Akrotiri, Thera.}
\label{fig:2}       
\end{figure}

\subsection{Choice of possible classes of potential prototype curves based on historical and archaeological grounds}
\label{sec:9}
The idea emerged among the authors that certain wall-paintings excavated at Mycenae, Thera and Crete might have been drawn by usage of prefabricated geometric stencils \cite{PapEx}, \cite{PapFrag}. A first step to verify this conjecture is the determination of a set of geometric shapes, whose conception and construction are not a priori prohibitive for the era, from an archaeological and historical point of view. For example, the linear spiral and the hyperbola can be constructed with the use of simple tools, even if this might require a considerable amount of novelty for the era. 

	Thus, extensive archaeological and historical analysis led to the conclusion that a set of geometric figures that could have been conceived and constructed by these civilizations, are the following: (a) Exponential spiral (b) The spiral generated by unwrapping a thread around a peg, usually called the involute of a circle (c) The linear or Archimedes' spiral (d) the ellipse (e) the parabola (f) the hyperbola.

	It is well known that spiral shapes appear in various prehistoric civilizations even centuries before the prehistoric Aegean civilizations. There are infinitely many types of spirals. Among them, the involute of a circle can be easily generated in everyday life events, while the exponential spiral can be found in various cockleshells. Thus, not surprisingly rough approximations of these 2 types of spirals are encountered quite early in various prehistoric civilizations. On the other hand, the linear spiral seemingly does not exist in nature. 
In Classical Ages, the conception of the linear spiral is so far attributed to Konon in the 3rd century B.C.. Next, in "On Spirals", Archimedes defines linear spiral and gives many fundamental properties and related theorems. 


\subsection{Determination of the stencils and the stencil parts most probably used for drawing "Lady of Mycenae" and the "Naked Boys"}
\label{sec:11}
In order to test if there are stencils obtained from the class of curves defined in Sect. \ref{sec:9} we first state the equations of these prototypes. Implicit functional forms of the adopted prototypes are given in Table \ref{tab:prot}. Analytical expressions for the flat curvature versions $K_S(x,y,a,b)$ that correspond to these prototypes families, are obtained straight from these functional forms.
\begin{table}
\caption{Descriptive functional forms of the adopted potential families of prototype curves}
\label{tab:prot}       
\begin{tabular}{ll}
\hline\noalign{\smallskip}
Model Curve & Functional Form \\
\noalign{\smallskip}\hline\noalign{\smallskip}
exponential & $f_{ES}(x,y|a,b) = x^2 +y^2 - a^2 \exp \left(2 b\theta (x,y) \right)$ \\spiral & $\theta(x,y)=\arctan \frac{y}{x} + (1-sgn(x))\frac{\pi}{2}$\\
involute of & $f_{IV}(x,y|a) = x^2 +y^2 - \alpha^2 \left(1+ \theta_{\epsilon}^2 (x,y) \right)$ \\a circle & $\theta_{\epsilon} + \arctan \theta_{\epsilon} =\arctan \frac{y}{x} + (1-sgn(x))\frac{\pi}{2}$\\
linear spiral & $f_{LS}(x,y|\kappa) = x^2 +y^2 - \kappa^2 \theta^2 (x,y)$ \\ & $\theta(x,y)=\arctan \frac{y}{x} + (1-sgn(x))\frac{\pi}{2}$\\
ellipse & $f_{EL}(x,y|a,b) = \left(\frac{x}{a}\right)^2 +\left(\frac{y}{b}\right)^2 - 1$\\
parabola & $f_{P}(x,y|a,b) = (ax+b)^2 - y$\\
hyperbola & $f_{H}(x,y|a,b) = \left(\frac{x}{a}\right)^2 -\left(\frac{y}{b}\right)^2 - 1$\\
\noalign{\smallskip}\hline
\end{tabular}
\end{table}

\begin{figure}[!t]
  \centerline{
  \includegraphics[width=0.24\textwidth]{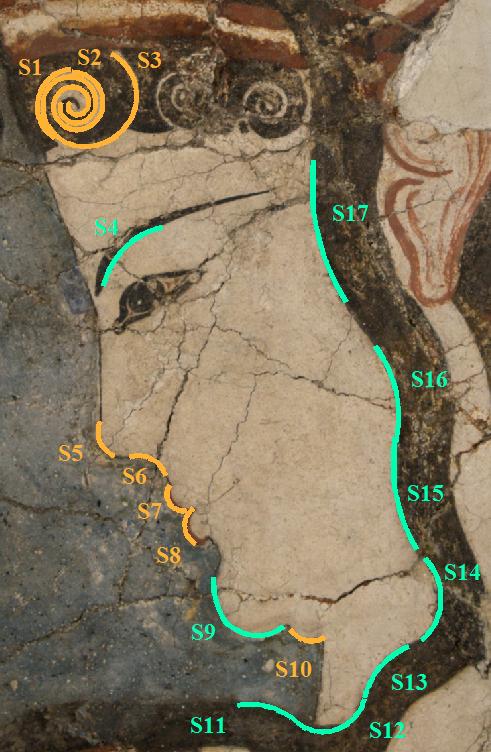}
  \includegraphics[width=0.24\textwidth]{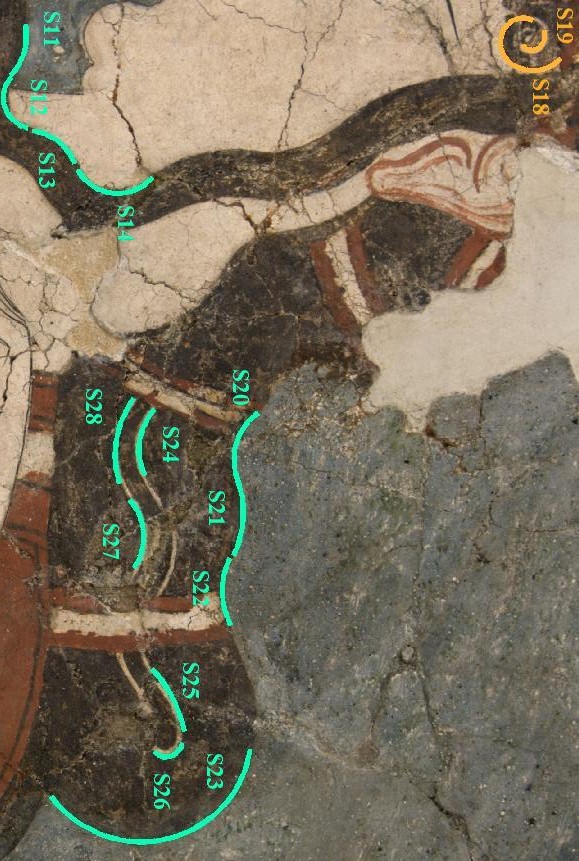}}
\caption{Optimal fitting of the stencil parts of the two linear spiral prototypes to the outlines of the "Lady of Mycenae" wall-painting. It seems that the two spirals were employed for drawing many of the contours of the face and hair details of the figure.}
\label{fig:4}       
\end{figure}

\begin{figure*}[!t]
\centerline{
  \includegraphics[width=0.4\textwidth]{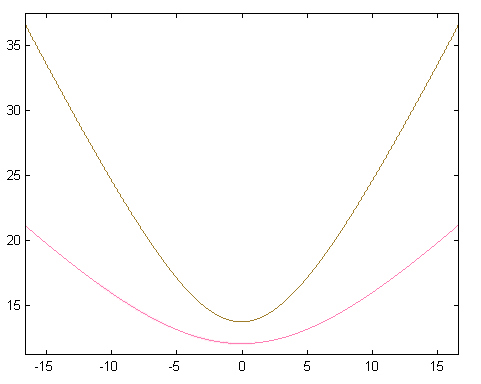}
  \includegraphics[width=0.4\textwidth]{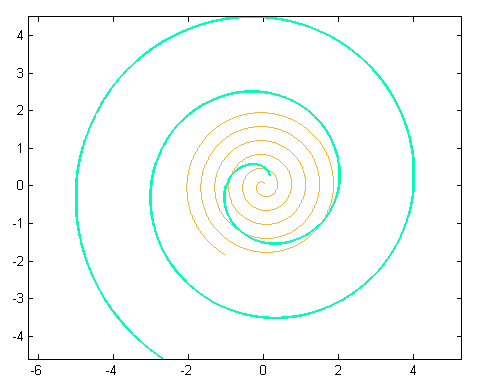}}
  \centerline{
  \includegraphics[width=0.81\textwidth]{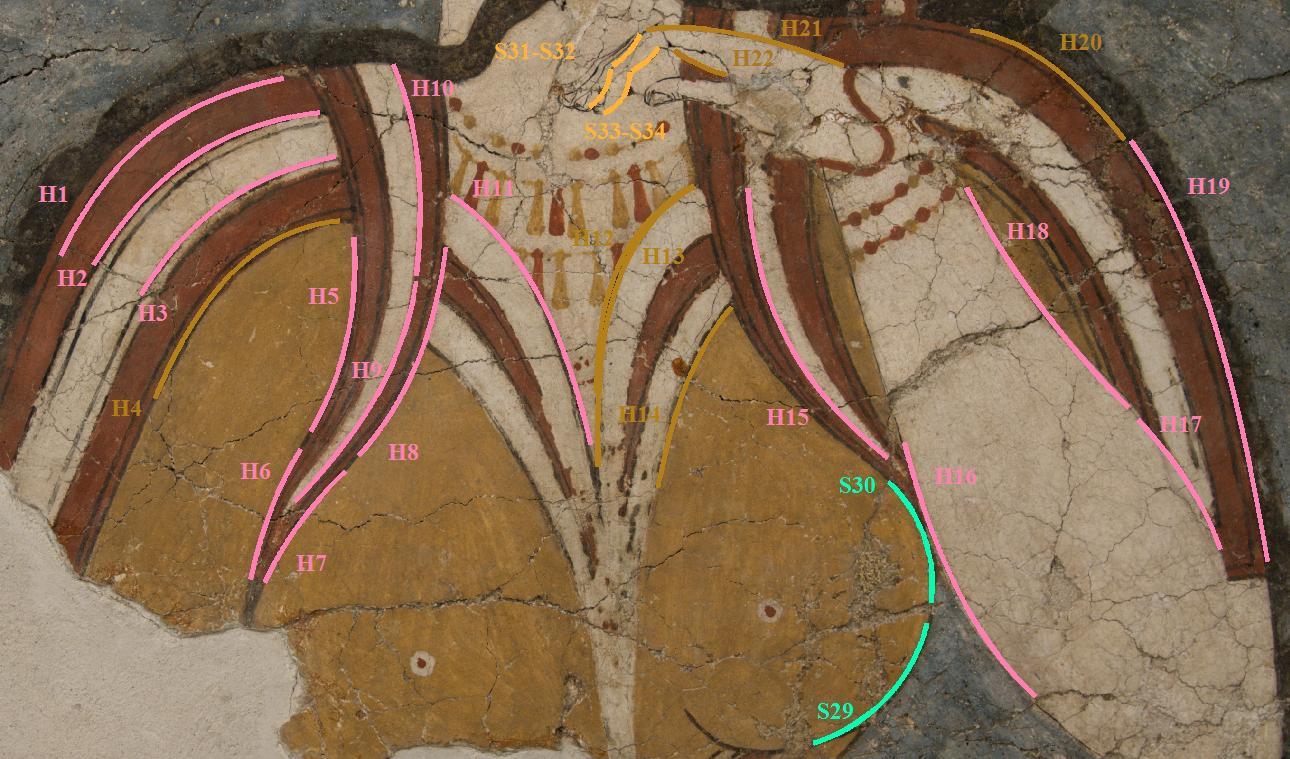}}
\caption{Demonstration of the excellent way, the geometric stencils described in Table \ref{tab:1} fit the corresponding object parts for the "Lady of Mycenae" wall-painting. Each color corresponds to a different stencil, as mentioned in Table \ref{tab:1}, and each determined stencil part is named in correspondence with Table \ref{tab:3}.}
\label{fig:5}       
\end{figure*}

\begin{figure*}[!t]
\centerline{
  \includegraphics[width=0.42\textwidth]{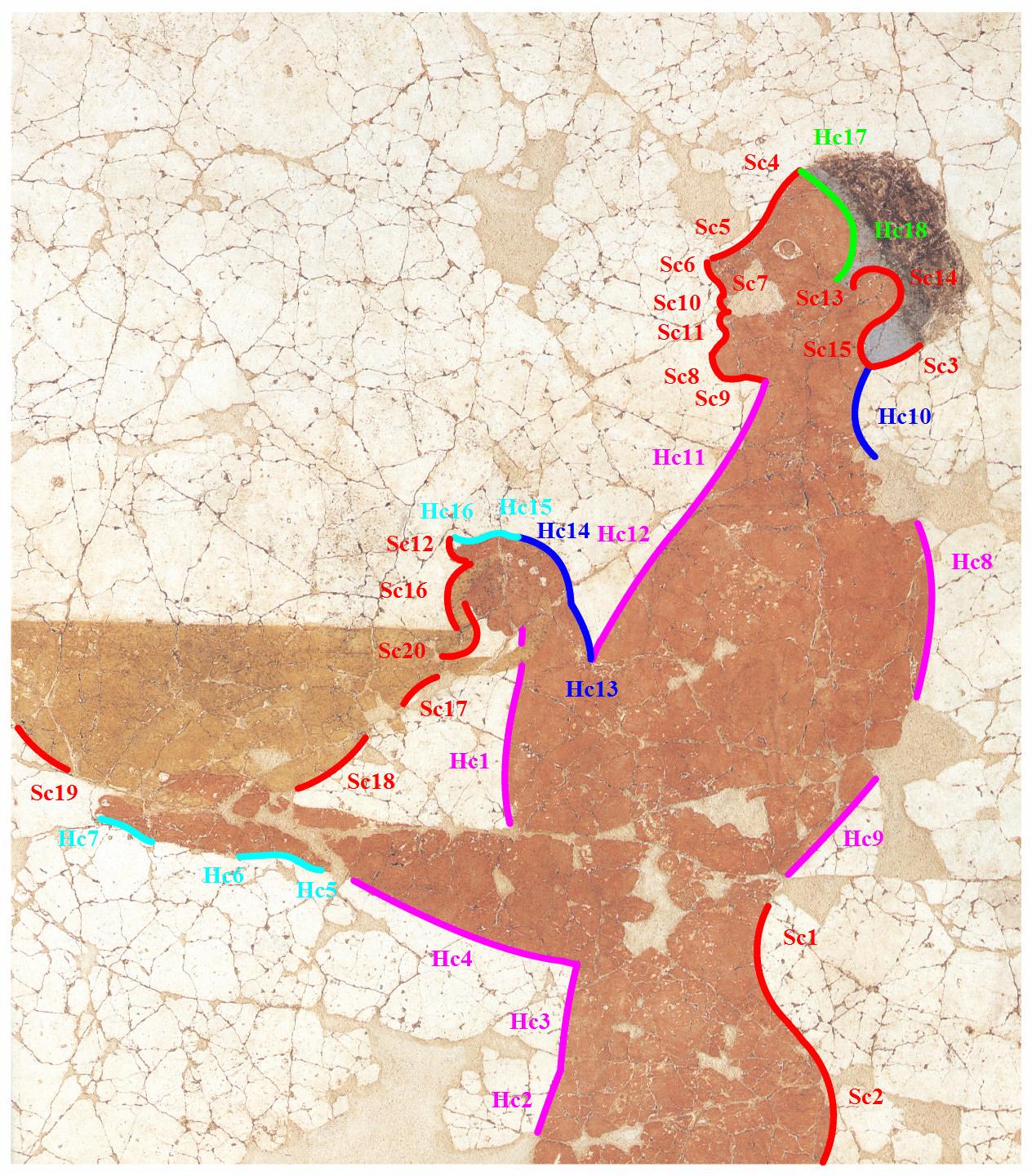}
  \includegraphics[width=0.42\textwidth]{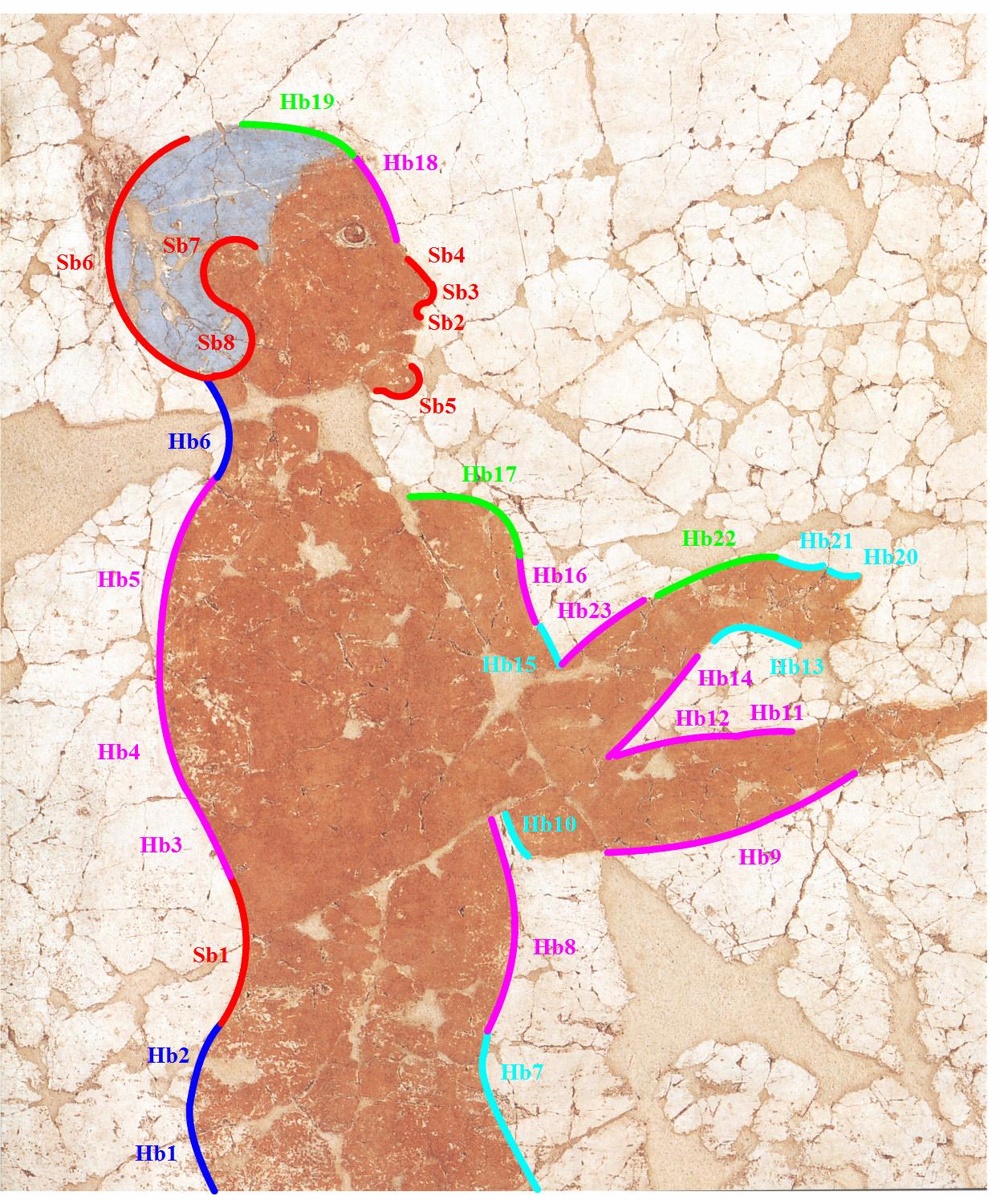}}
  \centerline{
  \includegraphics[width=0.38\textwidth]{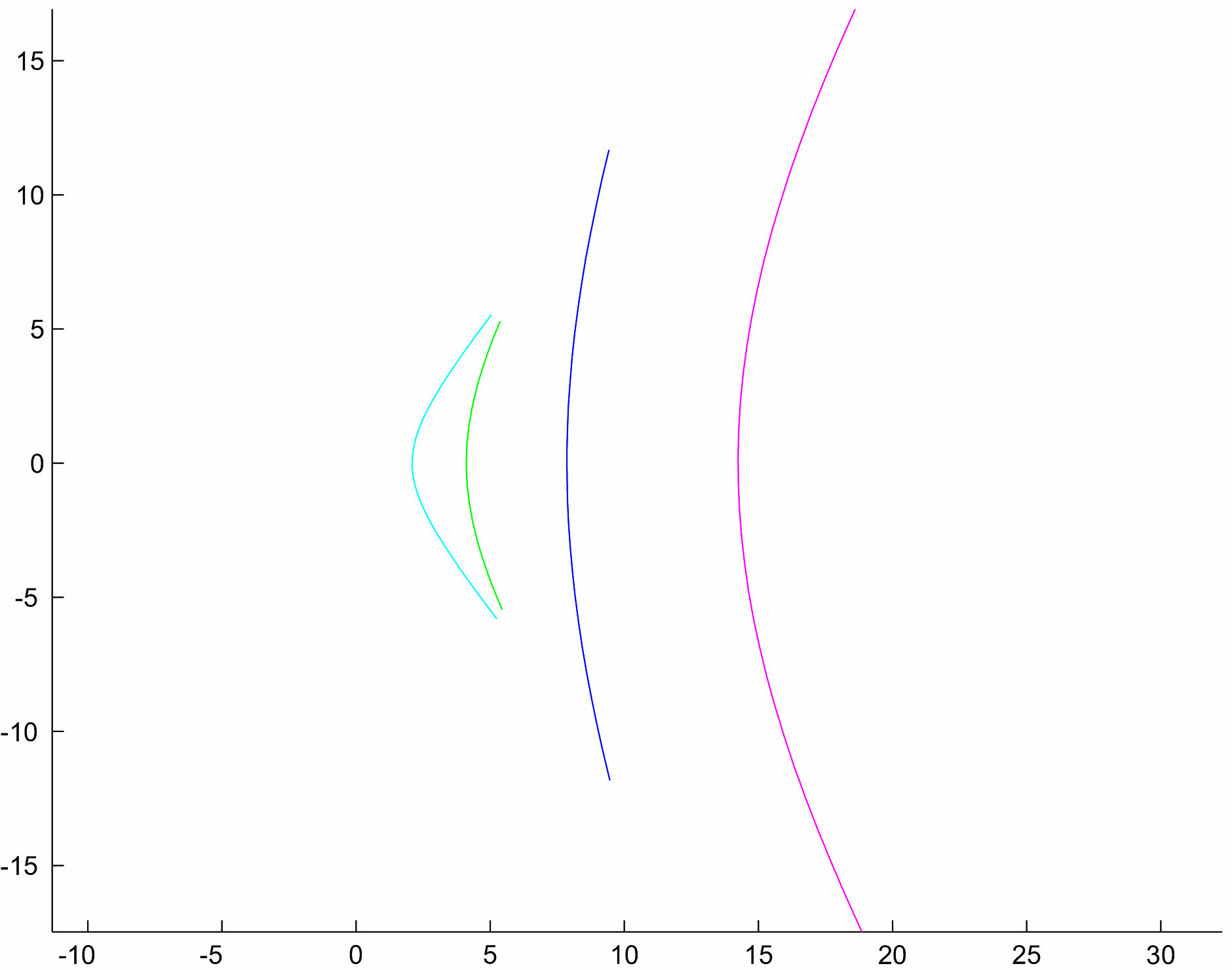}
  \includegraphics[width=0.38\textwidth]{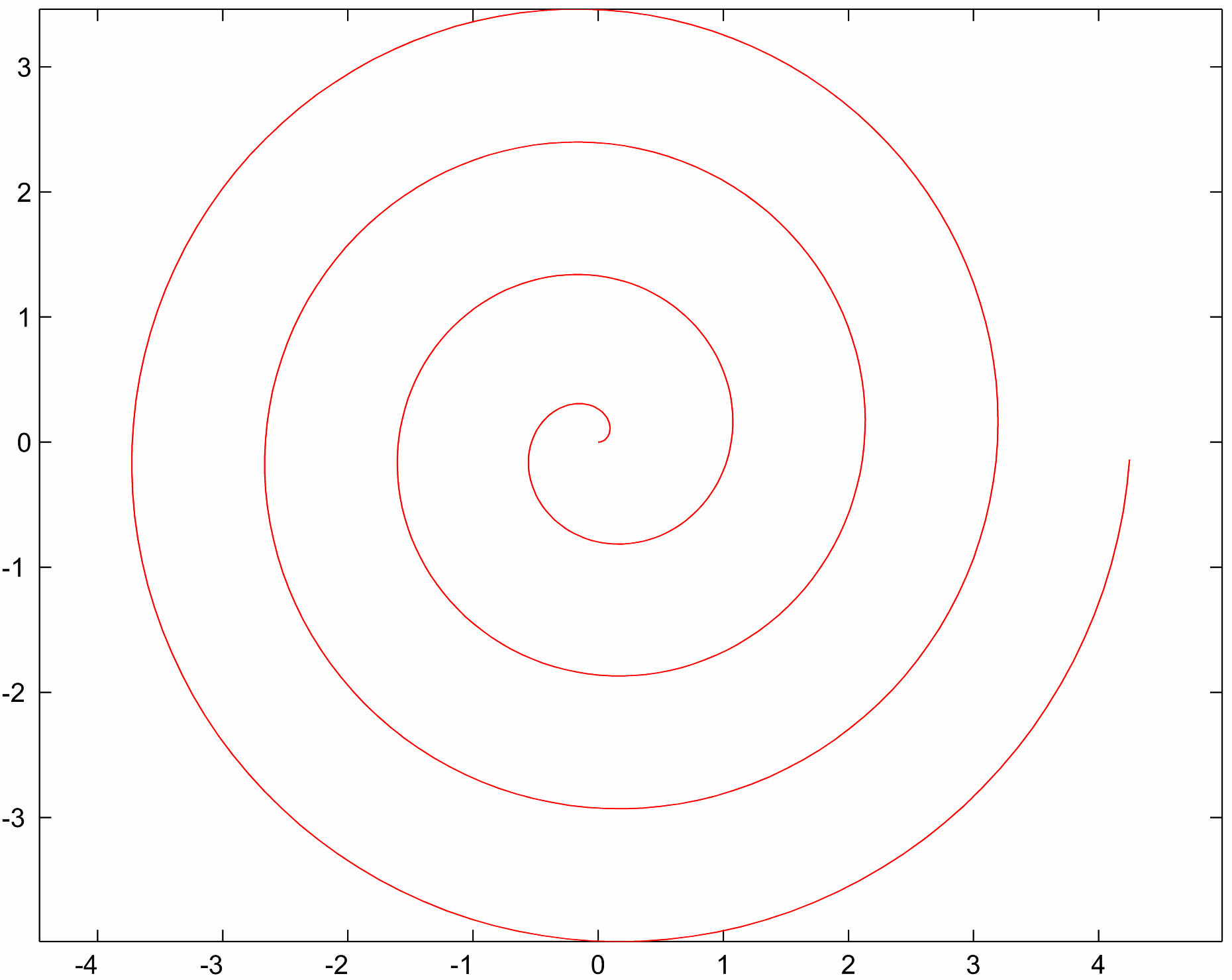}}
\caption{Manifestation of the excellent way, the geometric stencils described in Table \ref{tab:2} approximate the corresponding object parts for the "Naked Boys" wall-painting. Each color corresponds to a different stencil, as mentioned in Table \ref{tab:2}, and each determined stencil part is named in correspondence with Table \ref{tab:4}.}
\label{fig:6}       
\end{figure*}

We have applied the methodology presented in Sect. \ref{sec:3} to the prototype equations adopted in Table \ref{tab:prot} and for the main figures of the wall-paintings "Lady of Mycenae" and "Naked Boys".
First, photos of the considered wall paintings have been taken ("Lady of Mycenae") or scanned ("Naked Boys"). These images have been processed so as to extract the outlines of the painted figures. Next, using the methodology of Sect. \ref{sec:2}, objects and object parts of these outlines have been determined. The object parts with length greater than 1cm are employed so as to determine possible stencil parts used for creating these object parts. This is achieved by means of the procedure described in Sect. \ref{sec:5}, which offers the optimal prototype curve that best fits each available object part. After clustering the obtained stencil parts into groups of prototype curves with neighboring primary parameters values, as Sect. \ref{sec:6} describes, a unique prototype that optimally fits all object parts of the same cluster is determined, via the methodology described in Sect. \ref{sec:7}. We have reached the conclusion that both, "Lady of Mycenae" and "Naked Boys" wall paintings have been most probably drawn by means of linear spirals and hyperbolae stencils/guides. More specifically, it seems highly probable that one prototype linear spiral and four prototype hyperbolae had been used as guides for the drawing of the 
"Naked Boys" wall painting, while two linear spiral and two hyperbola prototypes have been used to draw the main parts of "Lady of Mycenae" outlines. Equivalently, all object parts of the contours of the main figures appearing in these wall-paintings impressively match to the prototypes specified in Table \ref{tab:1} for the "Lady of Mycenae" wall painting and in Table \ref{tab:2} for the "Naked Boys" wall paintings.
The stencil parts estimated for each object part are recorded in Tables \ref{tab:3} and \ref{tab:4} together with their lengths and their mean and maximum fitting errors. The visual results of fitting the stencil parts to paintings' object parts are presented in Figs. \ref{fig:4}, \ref{fig:5} and \ref{fig:6}. We point out, that for space economy reasons, the results concerning only two "Naked Boys" main figures are presented; very similar results hold for the other main figures, too.

The average matching error of each object part to a corresponding part of the prototype stencil, is always less than 0.39mm per pixel, while the maximum error is always less than 0.86mm.

\section{Conclusion}
\label{sec:12}
In this paper, we have tackled the problem that can be described as follows: Suppose that a set of outlines of drawn figures is given. Then, look for the minimum number of prototype functional forms that optimally fit these outlines, with a particularly small fitting error. During this fitting process, the outlines are partitioned in a number of contiguous parts of the maximum possible average length. To accomplish the aforementioned, we have shown that, in $\Re^2$, at most two parameters, we call primary, are sufficient to describe all curvature deformations of a twice continuously differentiable curve. Next, we employ the differentiable manifold formed by the two primary parameters and the position vectors of the considered functional forms. In this 4-manifold we perform exhaustive curvature driven optimization, to obtain both the optimal primary parameter values and the optimal relative placement of the prototype and drawn curves. We would like to point out that this fitting optimization behaves particularly well, even when the drawings have suffered serious wear. Moreover, the introduced methodology is of a more general applicability, for example in the case of fitting a prototype in noisy point clouds and in optimally matching particularly noisy contours.
 
Application of this methodology manifests that in the fresco technique which was widely spread in the Aegean civilizations before 1000 B.C., geometric guides were employed corresponding to linear (Archimedes) spirals and hyperbolae. It seems that this technique had been applied both in Akrotiri, Thera in the 17th century B.C. and in Mycenae in 14th-13th centuries B.C.. However, the parameters of the employed stencils differ, as it is demonstrated for the first time here. These results indicate perpetuated knowledge of constructing these geometric prototypes with impressive precision, as the particularly small matching error manifests. In addition, it is remarkable that these geometric conceptions appear with such an accuracy more than 1000 years before their axiomatic formulation and treatment by great mathematicians such as Archimedes, Euclid, Menaechmos, Appolonious, etc. One may perhaps speak for an emotional sense of geometry in these Late Bronze Age Civilizations, which was formally and axiomatically founded in the same region in the Classical Ages.

%

\begin{table}
\caption{Stencil Parameters For The "Lady Of Mycenae"}
\label{tab:1}       
\begin{tabular}{lll}
\hline\noalign{\smallskip}
Type of stencil & Color & Primary Parameters (cm)  \\
\noalign{\smallskip}\hline\noalign{\smallskip}
hyperbola 1 & pink & a=11.583  b=12.6025 \\
hyperbola 2 & brown & a=6.7500  b=13.7391 \\
linear spiral 1 & orange & k=0.0592 \\
linear spiral 2 & green & k=0.3175 \\
\noalign{\smallskip}\hline
\end{tabular}
\end{table}

\begin{table}
\caption{Stencil Parameters For The "Naked Boys"}
\label{tab:2}       
\begin{tabular}{lll}
\hline\noalign{\smallskip}
Type of stencil & Color in Figure 3 & Primary Parameters (cm)  \\
\noalign{\smallskip}\hline\noalign{\smallskip}
hyperbola 3 & magenta & a=14.24  b=20.12 \\
hyperbola 4 & light green & a=4.11  b=6.29 \\
hyperbola 5 & blue & a=7.86  b=17.63 \\
hyperbola 6 & cyan & a=2.09  b=2.52 \\
linear spiral 3 & red & k=0.169 \\
\noalign{\smallskip}\hline
\end{tabular}
\end{table}
\begin{table}
\caption{Stencil Parts Fitting Results for the "Lady of Mycenae"}
\label{tab:3}       
\begin{tabular}{lllll}
\hline\noalign{\smallskip}
Stencil & Type of & Mean & Max & Max \\
Part & Stencil & Error(cm) & Error(cm) & Length(cm) \\
\noalign{\smallskip}\hline\noalign{\smallskip}
H1-11,\\ 15-19,23 & hyperbola 1 & 0.0197 & 0.0555 & 16.6597 \\
H12-14,\\20-22,24,25 & hyperbola 2 & 0.0169 & 0.0449 & 9.9723 \\
S1-3,5-10,\\18,19,31-34 & spiral 1 & 0.0107 & 0.0272 & 5.2204 \\
S4,11-17,\\20-30 & spiral 2 & 0.0161 & 0.0374 & 9.1799 \\

\noalign{\smallskip}\hline
\end{tabular}
\end{table}

\begin{table}
\caption{Stencil Parts Fitting Results for the "Naked Boys"}
\label{tab:4}       
\begin{tabular}{lllll}
\hline\noalign{\smallskip}
Stencil & Type of & Mean & Max & Max \\
Part & Stencil & Error(cm) & Error(cm) & Length(cm) \\
\noalign{\smallskip}\hline\noalign{\smallskip}
Hb3-5,8,9,11,\\12,14,16,18,23\\
Hc1-4,8,9,11,12 & hyperbola 3 & 0.0209 & 0.0559 & 14.1541 \\
Hb1,2,6,\\ Hc10,13,14 & hyperbola 5 & 0.0168 & 0.0477 & 6.1036 \\
Hb7,10,13,\\15,20,21\\
Hc5-7,15,16 & hyperbola 6 & 0.0224 & 0.0470 & 7.7096 \\
Hb17,19,22\\ Hc17,18 & hyperbola 4 & 0.0193 & 0.0555 & 7.1760 \\
Sb1-8,Sc1-20 & spiral 3 & 0.0295 & 0.0497 & 15.5649 \\
\noalign{\smallskip}\hline
\end{tabular}
\end{table}

\appendices
\section{Minimization of the integral of absolute values via the integral of squared values}
\label{Ap1}
In this appendix, for the integrals $I_1(\Omega)=\int_{\Omega}|c(\omega)|d \omega$ and $I_2(\Omega)=\int_{\Omega}{c(\omega)}^2 d \omega$ used in Sect. \ref{sec:5_1} we will show that minimization of $I_2$ results minimization of $I_1$. 

In order to determine how minimization of $I_2$ affects minimization of $I_1$ we exploit the differentials $d I_1 = |c(\Omega)| d \Omega$, $d I_2 = c(\Omega) ^2 d \Omega = |c(\Omega)| d I_1$. The final relation, together with the one of $d I_1$, implies that the stationary points of $I_2$ are stationary points of $I_1$, even in the case where $c(\Omega)=0$ and that if there is a $\Omega^*$ such that $\forall \Omega$, $I_2(\Omega^*) \leq I_2(\Omega)$, only then $I_1(\Omega^*) \leq I_1(\Omega)$. Namely, if there is an $\Omega'$, for which $I_1(\Omega') \leq I_1(\Omega^*)$ and if $|c^M| = \underset{\Omega^* \rightarrow \Omega'}{\sup} |c(\Omega)|$, then relation $d I_2 = |c(\Omega)| d I_1$ offers $I_2(\Omega')-I_2(\Omega^*) \leq \int\limits_{\Omega^* \rightarrow \Omega'}c(\Omega) d I_1(\Omega) \leq |c^M| \left(I_1(\Omega')-I_1(\Omega^*)\right)\leq 0$, which violates $I_2(\Omega^*) \leq I_2(\Omega')$. Hence, relation $I_1(\Omega') \leq I_1(\Omega^*)$ does not hold.

\section{Minimization of $E_2(\partial \Omega)$ independently of primary parameters curve, $\alpha_S$, deformations}
\label{Ap2}
Let the descriptive equation of $S$ prototype curves family $f_S(\chi,\alpha)=f_c$ induce two curves, one on the $\chi=(x,y)$ sub-space $\chi_S(\xi, \alpha)$ and one on the $\alpha=(a,b)$ sub-space $\alpha_S(\beta, \chi)$. Then $\chi_S(\xi,\alpha_S(\beta,\chi))= \chi$, $\forall \xi$ and $\alpha_S(\beta,\chi_S(\xi,\alpha))= \alpha$, $\forall \beta$ implying that if $\chi \in \chi_S$ then $\partial_{\beta} \alpha = 0 \Leftrightarrow \alpha = \alpha_S(\beta_c) + \delta_{\alpha} \vec n_{\alpha}(\beta_c)$, $\beta_c$ is constant and if $\alpha \in \alpha_S$ then $\partial_{\xi}\chi = 0 \Leftrightarrow \chi = \chi_S(\xi_c) + \delta_{\chi} \vec n_{\chi}(\xi_c)$, $\xi_c$ is constant. Substituting the volume form \ref{eq:VlFrm} of $\partial \Omega$ in the integral (\ref{eq:int_curv2}) and separating integrals along $\chi_S$ and $\alpha_S$, $E_2$ reads
\begin{eqnarray}
 E_2 &=& \int\limits_{\beta_0}^{\beta_0+L_{\alpha}} \int\limits_{-M_{\alpha}(\beta_c)}^{M_{\alpha}(\beta_c)}\int\limits_{\xi_0}^{\xi_0+L_p}\left| \varepsilon_{\chi}(\xi,\beta_c,\delta_{\alpha}) \right|  d \xi d \delta_{\alpha} d \beta_c \nonumber \\
 &+& \int\limits_{\xi_0}^{\xi_0 + L_p} \int\limits_{0}^{\delta_{\chi}^{p}(\xi_c)}\int\limits_{\beta_0}^{\beta_0+L_{\alpha}}\left| \varepsilon_{\alpha}(\xi_c,\delta_{\chi},\beta)\right|  d \beta  d \delta_{\chi} d \xi_c \nonumber \\
 \label{eq:Lang}
\end{eqnarray}
where $\varepsilon_{\chi}$ and $\varepsilon_{\alpha}$ are defined via the formulas
\begin{eqnarray}
\varepsilon_{\chi}(\xi,\beta_c,\delta_{\alpha})&=&\left[K_S {c_{\chi}^{\alpha}}^2 \left(1+\frac{T_{\alpha}^2}{T_{\chi}^2}\right)\right]_{\vec r_{\xi}^{-}}^{\vec r_{\xi}^{+}} \nonumber \\ \varepsilon_{\alpha}(\xi_c,\delta_{\chi},\beta)&=&\left[K_S {c_{\chi}^{\alpha}}^2 \left(1+\frac{T_{\chi}^2}{T_{\alpha}^2}\right)\right]_{\vec r_{\alpha}^{-}}^{\vec r_{\alpha}^{+}}
\label{eq:epsilon}
\end{eqnarray}
and $\xi_c \in [\xi_0 , \xi_0+L_p]$, $\beta_c \in [\beta_0 , \beta_0+L_{\alpha}]$, $\delta_{\chi}^p(\xi_c)=(\vec r_S(\xi_c)-\vec r_p(\xi_c - \xi_0))^T \vec n_{\chi}$, $\vec r_{\xi}^{-}=\vec r_S (\xi ,\alpha(\beta_c,\delta_{\alpha}))$, $\vec r_{\xi}^{+}=\vec r_p(\xi - \xi_0,\alpha(\beta_c,\delta_{\alpha}))$, $\vec r_{\alpha}^{-}=(\chi(\xi_c, \delta_{\chi}) ,\alpha(\beta,-M_{\alpha}(\beta)))$, $\vec r_{\alpha}^{+}=(\chi(\xi_c, \delta_{\chi}) ,\alpha(\beta,M_{\alpha}(\beta)))$. If curves $\vec r_S$ and $\vec r_p$  optimally fit in the sense that their point by point distances are minimal, due to continuity of these curves there is at least one $\xi_c^{*} \in [\xi_0 , \xi_0+L_p]$ of zero distance between them, i.e. $\delta_{\chi}^p(\xi_c^{*}) = 0$. Then, formula (\ref{eq:Lang}) implies that there is a point $\alpha_S (\beta_c^{*}, x_S(\xi),y_S(\xi))$ of zero variance $\delta_{\alpha}(\beta_c^{*})$.  Moreover, consider all possible alignments between $\vec r_S$ and $\vec r_p$ that correspond to selecting any point of $\xi_c \in [\xi_0,\xi_0+L_p]$ and $\beta_c \in [\beta_0,\beta_0+L_{\alpha}]$, which satisfy $\xi_c^{*}$ and $\beta_c^{*}$ respectively. Then, the variance of $E_2$ at each one of these alignments is expressed as $\delta E_2 = \left. \frac{\partial E_2}{\partial (\delta_{\chi},\delta_{\alpha})}\right|_{\delta_{\alpha}=0 , \delta_{\chi}=0}$. The demand of zero variance of $E_2$ at its stationary points leads to equations \\
(a) $\int_{\beta_0}^{\beta_0+L_{\alpha}} \left. c_{\chi}^{\alpha}\frac{T_{\chi}}{T_{\alpha}}\right|_{\xi_c,\beta_c} \int_{\xi_0}^{\xi_0+L_p}\left| \varepsilon_{\chi}(\xi,\beta_c,0) \right|  d \xi d \beta_c =$ \\ $\int_{\xi_0}^{\xi_0 + L_p} \int_{\beta_0}^{\beta_0+L_{\alpha}}\left| \varepsilon_{\alpha}(\xi_c,0,\beta) \right|  d \beta  d \xi_c$ \\ (b) $\int_{\beta_0}^{\beta_0+L_{\alpha}} \int_{\xi_0}^{\xi_0+L_p}\left| \varepsilon_{\chi}(\xi,\beta_c,0) \right|  d \xi d \beta_c =$ \\ $\int_{\xi_0}^{\xi_0 + L_p} \left. c_{\chi}^{\alpha}\frac{T_{\alpha}}{T_{\chi}}\right|_{\xi_c,\beta_c} \int_{\beta_0}^{\beta_0+L_{\alpha}}\left| \varepsilon_{\alpha}(\xi_c,0,\beta) \right|  d \beta  d \xi_c$

If we consider the special case where the curve $M_{\alpha}(\beta)$ that bounds $\delta_{\alpha}$ is a dilated version of $a_S(\beta,\chi)$, then $\delta_{\alpha}$ is constant and, because of $\beta_c^{*}$, is equal to zero for all $\beta$. In this case $\varepsilon_{\alpha}(\xi_c,\beta)=0$ and the stationary points $(\xi_c,\beta)$ of $E_2$ are given by 
\begin{equation}
\int_{\beta_0}^{\beta_0+L_{\alpha}} \left. c_{\chi}^{\alpha}\frac{T_{\chi}}{T_{\alpha}}\right|_{\xi_c,\beta_c} \int_{\xi_0}^{\xi_0+L_p}\left| \varepsilon_{\chi}(\xi,\beta_c,0) \right|  d \xi d \beta_c = 0
 \label{eq:DLangIns}
 \end{equation}
Hence, expression (\ref{eq:Lang}) for $E_2$ is reformulated to $\delta E_2(\xi_0,\beta_0)$ $d \delta_{\alpha} =  d \delta_{\alpha} \int_{\beta_0}^{\beta_0+L_{\alpha}} \int_{\xi_0}^{\xi_0+L_p}\left| \varepsilon_{\chi}(\xi,\beta_c) \right|  d \xi d \beta_c$ and hence minimization of $E_2$ is achieved by minimization of $\delta E_2$. Then $(\xi_0^{*},\beta_0^{*})$ that minimize $\delta E_2$ read
\begin{equation}
(\xi_0^{*},\beta_0^{*}) = \arg \underset{(\xi_0,\beta_0)} \min \int_{\beta_0}^{\beta_0+L_{\alpha}} \int_{\xi_0}^{\xi_0+L_p}\left| \varepsilon_{\chi}(\xi,\beta_c,0) \right|  d \xi d \beta_c
\label{eq:MinLangIns}
\end{equation}

\section{Minimization of $E_2(\partial \Omega)$ over primary parameters curve, $\alpha_S$, deformations}
\label{Ap3}
As we mentioned in Sect. \ref{sec:7}, we look for a proper repositioning $\delta_{\chi}^C$ along the prototype curve's normals, where the object part and the prototype intersect, thus keeping $\xi_c(\delta_{\chi}^C)$ fixed, or equivalently $\xi_0(\delta_{\chi})=(\xi_0+L_p)(\delta_{\chi})=\xi_c(\delta_{\chi}^C)$. Looking at the fitting error (\ref{eq:E2N}), this re-estimation of the intersection points $\xi_c$ induces a selection $\delta_{\alpha}^C$ so as the level $\beta_c$ of the curvature deformation caused by the change of the parameters can be kept fixed, say equal to 0. This also corresponds to a selection of the proper $\alpha_p^* \in Q_S^C$. Among all these pairs of $\delta_{\chi}^C$ and $\delta_{\alpha}^C$,  we seek for those that zero the variation of the fitting error $\delta E_2^C(p) = \left. \frac{\partial E_2^C(p)}{\partial(\xi_c,\beta_c)}\right|_{\xi_c=\xi_0^*,\beta_c = 0}$. The demand $\delta E_2^C(p) = 0$ gives rise to the following equations \\
(a) $\int_{\delta_{\alpha}^C(\alpha_p^*)\in Q_S^C} \left. c_{\chi}^{\alpha} \frac{T_{\chi}}{T_{\alpha}}\right|_{\delta_{\chi}^C,\delta_{\alpha}^C} \int_{\xi_0}^{\xi_0+L_p}\left| \varepsilon_{\chi}^C(\xi,0,\delta_{\alpha}^C) \right|  d \xi d \delta_{\alpha}^C = $ \\ $\int_{m_{\chi}}^{M_{\chi}} \int_{\beta_0}^{\beta_0+L_{\alpha}}\left| \varepsilon_{\alpha}^C(\delta_{\chi}^C,\beta,0) \right|  d \beta  d \delta_{\chi}^C $ \\
(b) $\int_{m_{\chi}}^{M_{\chi}} \left. c_{\chi}^{\alpha}\frac{T_{\alpha}}{T_{\chi}}\right|_{\delta_{\chi}^C,\delta_{\alpha}^C}\int_{\beta_0}^{\beta_0+L_{\alpha}}\left|  \varepsilon_{\alpha}^C(\delta_{\chi}^C,\beta,0) \right|  d \beta  d \delta_{\chi}^C =$ \\ $\int_{\delta_{\alpha}^C(\alpha_p^*)\in Q_S^C} \int_{\xi_0}^{\xi_0+L_p}\left| \varepsilon_{\chi}^C(\xi,0,\delta_{\alpha}^C) \right|  d \xi d \delta_{\alpha}^C$ \\
where $\varepsilon_{\chi}^C$ and $\varepsilon_{\alpha}^C$ has been obtained by equation (\ref{eq:epsilon}) with boundaries defined in Sect. \ref{sec:7}.

But, as we have previously mentioned, the level $\beta$ of the prototype curvature deformation along the primary parameters level sets is kept fixed and equal to 0 and hence integral of $\varepsilon_{\alpha}$ vanishes letting the $E_2^C(p)$ stationary points equation be
\begin{equation}
\int_{\delta_{\alpha}^C(\alpha_p^*)\in Q_S^C} \left. c_{\chi}^{\alpha}\frac{T_{\chi}}{T_{\alpha}}\right|_{\delta_{\chi}^C,\delta_{\alpha}^C} \int_{\xi_0}^{\xi_0+L_p}\left| \varepsilon_{\chi}^C(\xi,0,\delta_{\alpha}^C) \right|  d \xi d \delta_{\alpha}^C = 0
 \end{equation}
 Then expression (\ref{eq:E2N}) for $E_2^C(p)$ is reformulated to 
\begin{equation*}
\delta E_2^C(p) d \beta_c =  d \beta_c \int_{\delta_{\alpha}^C(\alpha_p^*)\in Q_S^C} \int_{\xi_0}^{\xi_0+L_p}\left| \varepsilon_{\chi}^C(\xi,0,\delta_{\alpha}^C) \right| d \xi d \delta_{\alpha}^C
\end{equation*}
and hence minimization of $E_2^C(p)$ is achieved by minimization of $\delta E_2^C(p)$. Then $\xi_0^{*}(p)$ that minimizes $\delta E_2(p)$ reads
\begin{equation}
\xi_0^{*}(p) = \arg \underset{\xi_0} \min \int_{\delta_{\alpha}^C(\alpha_p^*)\in Q_S^C} \int_{\xi_0}^{\xi_0+L_p}\left| \varepsilon_{\chi}^C(\xi,0,\delta_{\alpha}^C) \right|  d \xi d \delta_{\alpha}^C
\end{equation}

\bibliographystyle{IEEEtran}
\bibliography{cit_mycenaea}   

\end{document}